\documentclass[format=acmsmall, review=false, screen=true]{acmart}

\usepackage{booktabs} 

\usepackage[ruled]{algorithm2e} 

\usepackage{amsmath}    
\usepackage{mdwlist}       

\renewcommand\footnotetextcopyrightpermission[1]{}  
\usepackage{mathtools}
\usepackage{multicol}
\usepackage{multirow}
\usepackage{tikz}
\usepackage{pgfplots}   	 
\usepackage{float} 
\usepackage{tcolorbox}   
\usepackage{listings}
\usepackage{amssymb}
\usepackage{filecontents}   
\usepackage{lipsum}            
\usetikzlibrary{arrows.meta}
\usepackage{amsmath,amssymb,amsthm}
\newtheorem{hyp}{Hypothesis}
\usepackage[noend]{algpseudocode}
\usepackage[ruled]{algorithm2e}

\newcommand*{\Comb}[2]{{}^{#1}C_{#2}}%

\newcommand{\system}{\texttt{DiffQue}}

\RequirePackage[textsize=scriptsize]{todonotes}

\SetAlFnt{\small}
\SetAlCapFnt{\small}
\SetAlCapNameFnt{\small}
\SetAlCapHSkip{0pt}
\IncMargin{-\parindent}

\acmJournal{TIST}
\acmVolume{9}
\acmNumber{4}
\acmArticle{39}
\acmYear{2010}
\acmMonth{3}
\copyrightyear{2009}

\setcopyright{none}

\begin{document}
\title[\system: Estimating Relative Difficulty of Questions]{\system: Estimating Relative Difficulty of Questions in Community Question Answering Services}

\author{Deepak Thukral }
\affiliation{%
  \institution{Amazon Pvt. Ltd}
  \country{India}}
\email{dthukral@amazon.com}

\author{Adesh Pandey}
\affiliation{%
 \institution{Tower Research Capital India Pvt. Ltd.}
 \country{India}}
\email{apandey@tower-research.com}

\author{Rishabh Gupta}
\affiliation{%
  \institution{MathWorks India Pvt Ltd}
  \country{India}
}
\email{rishabhg@mathworks.com}

\author{Vikram Goyal}
\affiliation{%
  \institution{IIIT-Delhi}
  \country{India}
}
\email{vikram@iiitd.ac.in}

\author{Tanmoy Chakraborty}
\affiliation{%
  \institution{IIIT-Delhi}
  \country{India}}
\email{tanmoy@iiitd.ac.in}

\begin{abstract}
Automatic estimation of relative difficulty of a pair of questions is an important and challenging problem in community question answering (CQA) services. There are limited studies which addressed this problem. Past studies mostly leveraged expertise of users answering the questions and barely considered other properties of CQA services such as metadata of users and posts, temporal information and textual content.  In this paper,  we propose \system, a novel system that maps this problem to a network-aided edge directionality prediction problem. \system~starts by constructing a novel network structure that captures different notions of difficulties among a pair of questions. It then measures the relative difficulty of two questions by predicting the direction of a (virtual) edge connecting these two questions in the network.  It leverages features extracted from the network structure, metadata of users/posts and textual description of questions and answers. Experiments on datasets obtained from two CQA sites (further divided into four datasets) with human annotated ground-truth show that \system~outperforms four state-of-the-art methods by a significant margin (28.77\% higher F\textsubscript{1} score and 28.72\% higher AUC than the best baseline). As opposed to the other baselines,  (i) \system~ appropriately responds to the training noise, (ii) \system~ is capable of adapting multiple domains (CQA datasets),
%
and (iii) \system~can efficiently handle `cold start' problem which may arise due to the lack of information for newly posted questions or newly arrived users.
\end{abstract}

%
%


%
%


\maketitle

\renewcommand{\shortauthors}{D. Thukral et al.}

\section{Introduction}
Programmers these days often rely on various community-powered platforms such as Stack Overflow, MathOverflow etc. -- also known as Community Question Answering (CQA) services to resolve their queries. A user posts a question/query which later receives multiple responses. The user can then choose the best answer (and mark it as `accepted answer') out of all the responses. Such platforms have recently gained huge attention due to various features such as quick response from the contributers, quick access to the experts of different topics, succinct explanation, etc.  For instance, in August 2010, Stack Overflow accommodated $300k$ users and $833k$ questions; these numbers have currently jumped to $8.3m$ users and $15m$ questions  posted\footnote{\url{https://stackexchange.com/sites/}}. This in turn provides tremendous opportunity to the researchers to consider such CQA services as large knowledge bases to solve various interesting problems  \cite{cscm4,msr-paper,chinese-paper}.  



Plenty of research has been conducted on CQA services, including  question search \cite{sigir1}, 
software development assistance \cite{rankingcrowdknowledge}, question recommendation \cite{recsys4}, etc. A significant amount of study is done in estimating user expertise \cite{sigir2}, recommending tags \cite{icsme1}, developing automated techniques to assist editing a post \cite{cscw1}, etc. Studies are also conducted to help developers mapping their queries to the required code snippets \cite{nlp2code}. 

\begin{figure}[!t]
\centering
\includegraphics[width=0.7\columnwidth]{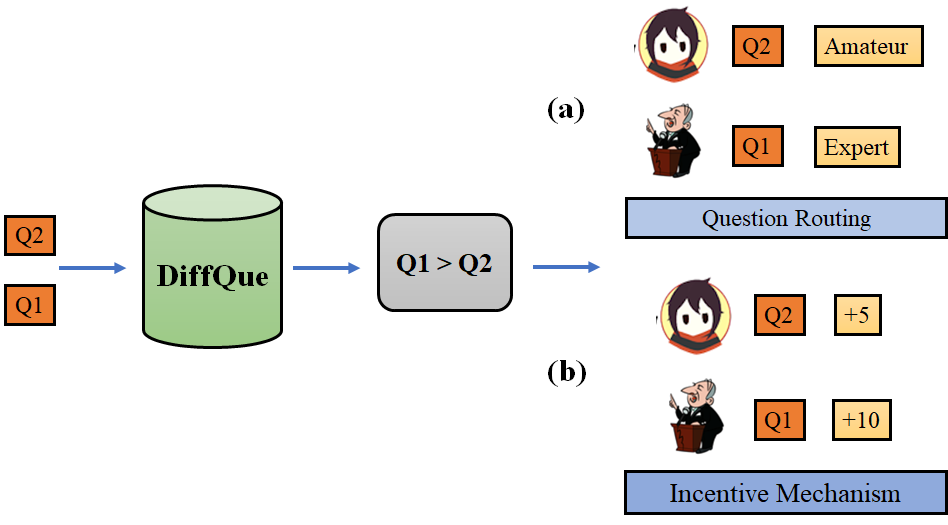}
\caption{A example showing how DiffQue can be utilised in different scenarios: (a) {\bf Question Routing}: Experts are presented with tougher questions as opposed to amateurs to save time and speed of the respective users; (b) {\bf Incentive Mechanism}:  If a user answers a tougher question, he should be provided more incentive in terms of reputation in the system.}\label{fig:motivation_example}
\end{figure}

{\bf Problem Definition and Motivation:} In this paper, {\em we attempt to automatically estimate the relative difficulty of a question among a given pair of questions posted on CQA services}. Such a system would help experts in retrieving questions with desired difficulty, hence making best use of time and knowledge. 
It can also assist to prepare a knowledge base of questions with varying level of difficulty. Academicians can use this system to set up their question papers on a particular topic. {\color{black} The proposed solution can be used in variety of scenarios; examples of such cases include: {\em question routing}, {\em incentive mechanism}, {\em linguistic analysis}, {\em analysing user behaviour}, etc. In question routing, questions are recommended  to the specified user as per his/her expertise. In incentive mechanism, point/reputation is allocated to the answerers depending upon the difficulty of question. Linguistic analysis allows to find similarity between language of question and its difficulty, and the significance of how it is framed and presented to audience. Analysing user behaviour identifies the preference of users in answering the questions, and the strategies used by them to increase their reputation in the system. We present a top example of two such use cases in Figure \ref{fig:motivation_example}.}
%
However, computing difficulty of a question is a challenging task as we cannot rely only on textual content or reputation of users. High reputation of a user does not always imply that his/her posted questions will be difficult.
Similarly, a question with embedded code (maybe with some obfuscation) does not always indicate the difficulty level. Therefore, we need to find a novel solution which uses different features of a question and interaction among various users to learn characteristics of the question and estimates the difficulty level. {\em To our knowledge, there are only three works} \cite{msr-paper,chinese-paper,regularised} {\em which tackled this problems before}. {\color{black} The major limitations of these methods are as follows: (i) they do not consider the varying difficulty of questions raised by same person; (ii) they often ignore low-level metadata; and (iii) none of them considered the temporal effect.}

{\bf Proposed Framework:} In this paper, we propose \system, a relative difficulty estimation system for a pair of questions that leverages a novel (directed and temporal) network structure generated from the user interactions through answering the posted questions on CQA services. \system~follows a two-stage framework (Figure \ref{fig:framework}). In the first stage, it constructs a network whose nodes correspond to the questions posted, and edges are formed  based on a set of novel hypotheses (which are statistically validated) derived from the temporal information and user interaction available in CQA services. These hypotheses capture different notion of `difficulties among a pair of questions'. In the second stage, \system~ maps the `relative question  difficulty estimation' problem to an `edge directionality prediction' problem. It extracts features from the user metadata, network structure and textual content, and runs a supervised classifier to estimate relatively difficult question among a question pair.  In short, \system~  systematically captures three fundamental properties of a CQA service -- user metadata, textual content and temporal information.

{\bf Summary of the Results:}
We evaluate the performance of \system~on two CQA platforms -- Stack Overflow (which is further divided into three parts based on time) and Mathematics Stack Exchange. We compare \system~with three other state-of-the-art techniques -- RCM \cite{regularised}, Trueskill \cite{msr-paper} and PageRank \cite{chinese-paper} along with another baseline (HITS) we propose here. All these baselines leverage some kind of network structure. Experimental results 
on human annotated data show that \system~outperforms all the baselines by a significant margin -- \system~achieves  $72.89\%$ F\textsubscript{1} score ({\em resp.} $72.9\%$ AUC) on average across all the datasets, which is $28.77\%$ ({\em resp.} $28.72\%$) higher than the best baseline. We statistically validate our network construction model and show that if other baselines had leveraged our network structure instead of their own, they could have improved the accuracy significantly (on average 2.67\%, 9.46\% and 17.84\% improvement for RCM, Trueskill and PageRank respectively;  however HITS originally leverages our network) compared to their proposed  system configuration. Further analysis on the performance of \system~reveals that  -- (i) \system~appropriately responds to the training noise, (ii) \system~is capable of adapting multiple domains, i.e., if it is trained on one CQA platform and tested on another CQA platform, the performance does not change much as compared to the training and testing on same domain, and (iii) \system~can handle cold start problem which may arise due to insufficient information of posts and users.

\begin{figure}[!t]
\centering
\includegraphics[width=0.8\columnwidth]{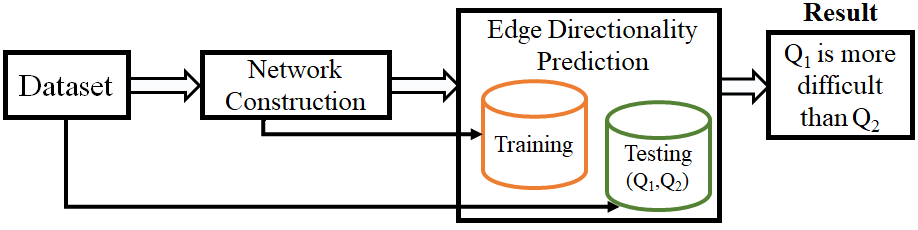}
\caption{A schematic architecture of \system.}\label{fig:framework}
\end{figure}

{\bf Contribution of the Paper:} In short, the major contributions of the paper are four-fold:
\begin{itemize}
\item We propose a novel network construction technique by leveraging the user interactions and temporal information available in CQA services. The network provides a relative ordering of pairs of questions based on the level of difficulty. We also show that the baselines could improve their performance if they  use our network instead of theirs. 

\item We map the problem of `relative difficulty estimation of questions' to an `edge directionality prediction' problem, which, to our knowledge, is the first attempt of this kind to solve this problem. Our proposed method utilizes three fundamental properties of  CQA services -- user information, temporal information and textual content.

\item \system~turns out to be superior to the state-of-the-art methods -- it not only beats the other baselines in terms of accuracy, but also appropriately responds to the training noise and handles cold start problem.  

\item As a by-product of the study, we generated huge CQA datasets and manually annotated a set of question pairs based on the difficult level. This may become a valuable resource for the research community.

\end{itemize}
{\bf For the sake of reproducible research, we have made the code and dataset public at \url{https://github.com/LCS2-IIITD/DiffQue-TIST}}. 

\section{Related Work}
Recently, we have seen expansion of various CQA services, including Stack Overflow, Quora, Reddit etc. There are number of studies that involve these CQA services. Here we organize the literature review in two parts: general study involving CQA and studies on difficulty ranking of questions.

\subsection{{\color{black}General Study Involving CQA and Ranking mechanism}}
CQA has grown tremendously in size, thus providing a huge databases to the research community. With such active contribution from the community, the data can be used for many important problems. \citet{codeexample} mined the Stack Overflow data and provided ways to identify helpful answers. They found characteristics of good code example and different attributes that should accompany the code to make it more understandable.
\citet{Chowdhury:2019} released CQASUMM, the first annotated CQA summarization dataset, and developed VIZ-Wiki, a browser extension to summarize the threads in CQA sites \cite{chowdhury2018viz}.

\citet{recomm} recommended questions available in CQA to experts based on their domain expertise. They find top $K$ potential experts for a question in the CQA service. \citet{kdd1} observed user's lifetime in CQA services and found that it does not follow  exponential distributions. There are many studies that analyzed temporal effect in CQA sites \cite{ieee1,wsdm1,icwsm2}. \citet{icwsm2} showed that there are two strategies to develop online authority: progressively developing reputation or by using prior acquired fame. \citet{ieee1} found online user activities time-dependent, and proposed question routing mechanism for timely answer. \citet{wsdm1} developed K-SC algorithm on Twitter dataset and showed that online content exhibits six temporal shapes. 
\citet{temporaljournal} found three parameters to capture temporal significance and to personalize PageRank -- frequency factor (trustworthy users are often active), built-up time length factor (the longer the time between registration of user and link creation, the more trustworthy the user) and similarity factor (the pattern based on which two users add links). Many efforts were made  in the field of expert identification, which mostly focus on link analysis and latent topic modeling techniques \cite{cikmm}. \citet{kdd08} focused on best answers provided by users to quantify experts. It is based on the concept of in-degree. \citet{wsdmtwit} combined topical similarity with link analysis for expert identification.  Some techniques leveraged Gaussian mixture models to solve the problem \cite{rel25, rel26}. \citet{cikm05} attempted to find similar questions and recommend them by exploring semantic features.

{\color{black}
Appropriate Ranking of entities (documents, search results, recommended items) has been one of the major concerns in information retrieval and data mining communities. Pioneering attempts were made in late 90', which resulted several query-independent ranking schemes such as PageRank \cite{pagerank}, TrustRank \cite{gyongyi2004combating}, BrowseRank \cite{Liu:browserank}, etc. The limitations of these ranking mechanisms were further addressed in several query-dependent ranking models such as HITS \cite{kleinberg1999authoritative}, SALSA \cite{lempel2001salsa}, LSI \cite{deerwester1990indexing}, vector space model \cite{baeza2011modern}, statistical language model \cite{zhai2008statistical},  etc. The problem in such conventional ranking models is that the model parameters are usually hard to tune manually, which may sometime lead to overfitting. Moreover, it is non-trivial to combine the large number of conventional models proposed in the literature to obtain an even more effective model. To overcome these problem, ``learning-to-rank'' was introduced  that uses
machine learning technologies to solve the problem of ranking \cite{liu2009learning}. Methods dealing with learning-to-rank can be categorized into three classes \cite{cao2007learning} -- pair-wise \cite{chu2005gaussian,chu2005preference}, point-wise \cite{qin2007ranking,tsai2007frank}, and list-wise \cite{burges2007learning,cao2007learning}. A number of machine learning algorithms have been proposed which optimize relaxations to minimize the number of pairwise misordering in the ranking produced. Examples include RankNet \cite{burges2005learning}, RankRLS \cite{pahikkala2009efficient}, RankSVM \cite{lee2014large,joachims2002optimizing,herbrich1999support}, etc.
\citet{freund2003efficient} proposed RankBoost, an adaptation of boosting mechanism for learning rank. \citet{de2019reinforcement} used reinforcement learning for ranking and recommendation. Learning-to-rank is also used in many applications -- specification mining \cite{cao2018rule}, named-entity recognition \cite{nguyen2018disease}, crowd counting \cite{liu2018leveraging}, etc. Machine learning community has also witnessed an adaptation of deep learning techniques for learning-to-rank \cite{cheng2018extreme,he2018hashing,sasaki2018cross}. Nevertheless, we here consider RankSVM and RankBoost as two basedlines for \system.  

}

\subsection{Difficulty Ranking of Questions}\label{sec:rw:qd}
{\color{black} \citet{hassansoqde} developed SOQDE, a supervised learning based model and performed an empirical study to determine how the difficulty of a question impacts its outcome. \citet{arumae2018study} used variable-length context convolutional neural network model to investigate whether it is possible to predict if question will be answered. \citet{chang2013routing} developed a routing mechanism that uses availability and expertise of the users to recommend answerers to a question.} \citet{chinese-paper} and \citet{msr-paper} attempted to solve the problem of finding the difficulty of tasks taking the underlying network structure into account. Ranking of questions according to the difficulty is a complex modeling task and highly subjective. 
\citet{chinese-paper} constructed a network in such a way that if user $A$ answers questions $Q_1$ and $Q_2$, among which only the answer of $Q_1$ is accepted, then there will be a  directed edge from $Q_2$ to $Q_1$. Following this, they used {\bf PageRank} as a proxy of the difficulty of a question.
\citet{msr-paper} created a user network where each question $Q$ is treated as a pseudo user U$_{Q}$. It considers four competitions: one between pseudo user U$_{Q}$ and asker U$_{a}$, one between pseudo user U$_{Q}$ and the best answerer U$_{b}$, one between the best answerer U$_{b}$ and asker U$_{a}$ and one between the best answerer U$_{b}$ and each of the non-best answerers of the question $Q$. It then uses {\bf Trueskill}, a Bayesian skill rating model. Both \citet{chinese-paper} and \citet{msr-paper} used neither textual content nor temporal effect, which play a crucial role in determining the importance of questions over time. \citet{regularised} used the same graph structure as \cite{msr-paper} and proposed Regularized Competition Model ({\bf RCM}) to capture the significance of difficulty. It forms $\theta \in \mathcal{R}^{M+N}$, denoting the `expert score' of pseudo users -- initial $M$ entries are expertise of users while further $N$ are difficulty of questions. For each of the competitions, $x_k$ vector is formed where ${x_i}\textsuperscript{k}$ = 1, ${x_j}\textsuperscript{k}$ = -1 and $y_k$ = 1 if $i$ wins over $j$, else $y_k$ = -1. The algorithm starts at initial $\theta$ and proceeds towards negative subgradient, $\theta_{t+1}$ = $\theta_t$ - $\gamma_t$*$\triangledown$$\mathcal{L}$($\theta_t$), where $\triangledown$$\mathcal{L}$($\theta_t$) is the subgradient and $\gamma_t$ is set as 0.001. Further, classification of examination questions according to Bloom's Taxonomy \cite{rccst1} tried to form an intuition of difficulty of questions; but it is not very helpful in case of CQA services where there are many users, and individual users have their own ways of expressing the questions, or sometimes there is code attached with question where normal string matching can trivialize the problem. We consider three methods \cite{msr-paper,chinese-paper,regularised} mentioned above as baselines for \system.

{\color{black} {\bf How \system\ differs from others:} \citet{chinese-paper} ran PageRank on a network based on a single hypothesis that if user wins in task X and not in Y, then there is directed edge from X to Y. They used PageRank to compute the hardness of questions. 
In their method, PageRank can also be replaced by authoritativeness \cite{hits} to rank the questions (which we also treat as a baseline). Both of these do not take into account some important points such as varying of difficulty of questions asked by the same person, or different persons who answered their questions. \citet{msr-paper} and \citet{regularised} used the same network, and applied Bayesian network and regularisation respectively. \citet{msr-paper} did not consider cold start problem and textual content, while \citet{regularised} did not take low-level features such as accepted answers, time difference between posting and acceptance, etc. The temporal effect which plays a significant role in this type of setting is also not discussed in the previous works. \system\ takes into account all the limitations/disadvantages of the previous works mentioned above.}


\section{Datasets} \label{sec:dataset}

We collected questions and answers from two different CQA services -- (i) Stack Overflow\footnote{\url{https://stackoverflow.com/}} (SO) and Mathematics Stack  Exchange\footnote{\url{https://math.stackexchange.com/}} (MSE), both of which are extensively used  by programmers or mathematicians to get their queries resolved. 
A brief description of the datasets are presented below (see Table \ref{tab:dataset}). 
{\color{black} We chose SO because it is the most preferable platform for programmers to resolve queries and has a very active community. Such a high level of engagement among users would help in modeling the underlying network better. Similarly, MSE is widely used by mathematicians and has an active community. Most importantly, both the datasets are periodically archived for research purposes. }


\subsection{Stack Overflow (SO) Dataset}
This dataset contains all the questions and the answers of Stack Overflow till Aug'17, accommodating more than $10m$ questions (70\% of which are answered) and more than $20m$ answers in total. In this work, we only considered `Java-related' questions as suggested in \cite{msr-paper}. Previous work \cite{msr-paper}  considered all questions and answers submitted to SO during Aug'08-Dec'10 (referred as {\bf SO1}). Along with this dataset, we also considered two more datasets -- we collected two recent datasets from SO posted in two different time durations -- questions and answers posted in Jan'12 -- Dec'13 (referred as {\bf SO2})  and in Aug'15 -- Aug'17 (referred as {\bf SO3}). Analyzing datasets posted in different time points can also help in capturing the dynamics of user interaction in different time points, which may change over time. User metadata was also available along with the questions and answers.  

\subsection{Mathematics Stack Exchange (MSE) Dataset}
This dataset contains all the questions posted on MSE till Aug'17 and their answers. There are about $800k$ questions and about a million answers. About 75\% questions are answered. Here we only extracted questions related to the following topics: Probability, Permutation, Inclusion-Exclusion and Combination\footnote{The topics were chosen based on the expertise of the human annotators who further annotated the datasets.}. The questions and answers belonging to these topics are filtered from the entire database to prepare our dataset. The number of users in the chosen sample is $47,470$. Unlike SO, we did not divide MSE into different parts because total number of questions present in MSE is less compared to SO (less than even a single part of SO). Further division might have reduced the size of the dataset significantly. 

Both these datasets are available as dump in xml format. Each post (question/answer) comes with other metadata: upvotes/downvotes, posting time, answer accepted or not, tag(s) specifying the topics of the question etc. Similarly, the metadata of users include account id, registration time and reputation. Unlike other baselines \cite{msr-paper,chinese-paper,regularised}, we also utilize these attributes in \system. 



\begin{table}[!t]
  \centering
    \begin{tabular}{c|c|c|c|c}
     \hline
 {\bf Dataset} & {\bf \# questions} & {\bf \# answers} & {\bf \# users} & {\bf Time period} \\
 \hline
 SO1 & 100,000 & 289,702 & 60,443 & Aug'08 --  Dec'10 \\ 

 SO2 &  342,450  & 603,402 &  179,827 & Jan'12 -- Dec'13 \\
 SO3 & 440,464 & 535,416 & 274,421 & Aug'15 -- Aug'17 \\\hline

 MSE & 92,686 & 119,754 & 47,470 & July'10 -- Aug'17  \\ 
 \hline
  \end{tabular}
      \caption{{\color{black}Statistics of the datasets used in our experiments (Java-related questions for SO and Probability, Permutation, Inclusion-Exclusion and Combination related questions for MSE).}}
      \label{tab:dataset}
      \vspace{-3mm}
      \end{table}
\section{\system: Our Proposed Framework}
\system~first maps a given CQA data to a directed and longitudinal\footnote{Nodes are arranged based on their creation time.} network where each node corresponds to a question, and an edge pointing from one question to another question indicates that the latter question is harder than the former one. Once the network is constructed, \system~ trains an edge directionality prediction model on the given network and predicts the directionality of a virtual edge connecting two given questions under inspection. The framework of \system~is shown in Figure \ref{fig:framework}. Rest of the section elaborates individual components of \system.   

\vspace{1mm}
\noindent\fbox{%
    \parbox{\columnwidth}{%
{$\bullet$ \bf Nomenclature:} Throughout the paper, we will assume that {\bf Bob} has correctly answered  {\bf Robin}'s question on  a certain topic, and therefore Bob has more expertise than Robin on that topic.\\
{$\bullet$ \bf Assumption:} Whenever we mention that Bob has more expertise than Robin, it is always w.r.t. a certain topic, which may not apply for other topics.

}}\vspace{2mm}

\subsection{Network Construction}\label{sec:networkconst}
\system~models the entire dataset as a directed and longitudinal network $G=(V,E)$, where $V$ indicates a set of vertices and each vertex corresponds to a question; $E$ is a set of edges. Each edge can be of one of the following three types mentioned below.

\noindent {\bf \underline{Edge Type 1:}} An expert on a certain topic does not post trivial questions on CQA sites. Moreover, s/he answers those questions which s/he has expertise on. We capture these two notions in Hypothesis \ref{hyp1}. 

\begin{hyp}\label{hyp1}
If Bob correctly answers question $Q$ asked by Robin on a certain topic, then the questions asked by Bob later on that topic will be considered more difficult than $Q$.
\end{hyp}
The `correctness' of an answer is determined by the `acceptance' status of the answer or positive number of upvotes provided by the users (available in CQA services).

Let $Q_R$ be the question related to topic $\Upsilon$ Robin posted at time $T_R$ and Bob answered $Q_R$. Bob later asked $n$ questions related to $\Upsilon$, namely $Q_{B_1}, Q_{B_2}, \cdots, Q_{B_n}$ at $T_{B_1}, T_{B_2}, \cdots, T_{B_n} (>T_R)$ respectively.  Then the difficulty level of each $Q_{B_i}$, denoted by $diff(Q_{B_i})$ will be more than that of  $Q_{R}$, i.e., $diff(Q_{B_i})\geq diff(Q_{R}), \forall i$. The intuition behind this hypothesis is as follows -- since Bob correctly answered $Q_R$, Bob is assumed to have more expertise than the expertise required to answer $Q_R$. Therefore, the questions that Bob will ask later from topic $\Upsilon$ may need more expertise than that of  $Q_R$. 

We use this hypothesis to draw edges from easy questions to difficult questions as follows: an edge $e=\langle x,y \rangle \in E$ of type 1 indicates that $y$ is more difficult than $x$ according to Hypothesis \ref{hyp1}. Moreover, each such edge will always be a forward edge, i.e., a question asked earlier  will point to the question asked later. Edge $\langle Q_{R_2}, Q_{B_3} \rangle$ in Figure \ref{fig:example} is of type 1.

One may argue that the {\em answering time} is also important in determining type 1 edges -- if at $T_1$ Bob answers $Q_R$ which has been asked by Robin at $T_{Q_R}$ (where $T_1>T_{Q_R}$), we should consider all Bob's questions posted after $T_{1}$ (rather than $T_{Q_R}$) to be difficult than $Q_R$. However, in this paper we do not consider  answering time separately and assume question and answering times to be the same because the time difference between posting the question and correctly answering the  question, i.e.,  $T_1-T_{Q_R}$ seems to be negligible across the datasets (on average $7.21$, $1.33$, $7.12$, $1.34$ days for SO1, SO2, SO3 and MSE respectively). \\

\noindent \textbf{\underline{Edge Type 2:}} It is worth noting that an edge of type 1 only assumes Bob's questions to be difficult which {\em will be} posted later. It does not take into account the fact that all Bob's contemporary questions (posted very recently in the past on the same topic) may be difficult than Robin's current question; even if the former questions may be posted slightly before the latter question. We capture this notion in Hypothesis \ref{hyp2} and draw type 2 edges.

\begin{hyp}\label{hyp2}
If Bob correctly answers Robin's question $Q$ related to topic $\Upsilon$, then Bob's very recent posted questions on $\Upsilon$ will be more difficult than $Q$.
\end{hyp}

Let $Q_{B_1}, Q_{B_2}, \cdots$ be the questions posted by Bob at $T_{B_1}, T_{B_2}, \cdots$ respectively. Bob has answered Robin's question $Q_R$ posted at $T_{R}$ and $T_{R}\geq T_{B_i}, \forall i$. However, the difference between $T_{R}$ and $T_{B_i}$ is significantly less, i.e., $T_R-T_{B_i}\leq \delta_t, \forall i$ which indicates that all these questions are contemporary. According to Hypothesis \ref{hyp2}, $diff(Q_{B_i})\geq diff(Q_R)$.

We use this hypothesis to draw edges from easy questions to difficult questions as follows: an edge $e=\langle x,y \rangle \in E$  of type 2 indicates that $y$ is more difficult than $x$, and $y$ was posted within $\delta_t$ times before the posting of $x$. Note that each such edge will be a backward edge, i.e., a question asked later may point to the question asked earlier. Edge $\langle Q_{R_2}, Q_{B_2} \rangle$ in Figure \ref{fig:example} is of type 2. \\

\noindent{\bf \underline{Edge Type 3:}} We further consider questions which are posted by a single user on a certain topic over time, and propose Hypothesis \ref{hyp3}. 

\begin{hyp}\label{hyp3}
A user's expertise on a topic will increase over time, and thus the questions that s/he will ask in future related to that topic will keep becoming difficult. 
\end{hyp}

Let $Q_{B_1}, Q_{B_2}, \cdots$ be the questions posted by Bob\footnote{Same hypothesis can be applied to any user (Bob/Robin).} at $T_{B_1}, T_{B_2}, \cdots$ , where $T_{B_{i+1}}> T_{B_i}, \forall i$. Then $diff(Q_{B_{i+1}})>diff(Q_{B_i}), \forall i$. The underlying idea is that as the time progresses, a user gradually becomes more efficient and acquires more expertise on a particular topic. Therefore, it is more likely that s/he will post questions from that topic which will be more difficult than his/her previous questions. 

We use this hypothesis to draw an edge from easy questions to difficult questions as follows: an edge $e=\langle x,y \rangle \in E$ of type 3 indicates that: (i) both $x$ and $y$ were posted by the same user, (ii) $y$ was posted after $x$ and there was no question posted by the user in between the posting of $x$ and $y$ (i.e., $T_y>T_x \& \nexists z: T_y>T_z>T_x$), and (iii) $y$ is more difficult than $x$. Note that each such edge will also be a forward edge. Edge $\langle Q_{B_2}, Q_{B_3} \rangle$ in Figure \ref{fig:example} is of type 3, but $Q_{B_2}$ and $Q_{B_4}$ are not connected according to this hypothesis.

Note that an edge can be formed by more than one of these hypotheses. However, we only keep one instance of such edge in the final network (and avoid multi-edge network). In Section \ref{sec:hypothesis}, we will show that all these hypotheses are statistically significant. The number of edges of each type is shown in Table \ref{edges}. {\color{black} Note that the hypotheses proposed in \cite{msr-paper} are different from those we proposed here. The earlier work did not consider the temporal dimension at all. They compared question asker, question, best  answerer, and all others who answered the question. Further their hypotheses related question, answerer and other answerer only within a thread and not across different threads. On the other hand, hypothesis 1 proposed in this paper establishes a relationship between the questions that are answered and all questions asked by the same answerer in future. Our proposed hypothesis 2 establishes a relationship between recently asked questions by the answerer and the question answered by the answerer. }

\begin{figure}[!t]
    \centering
           \includegraphics{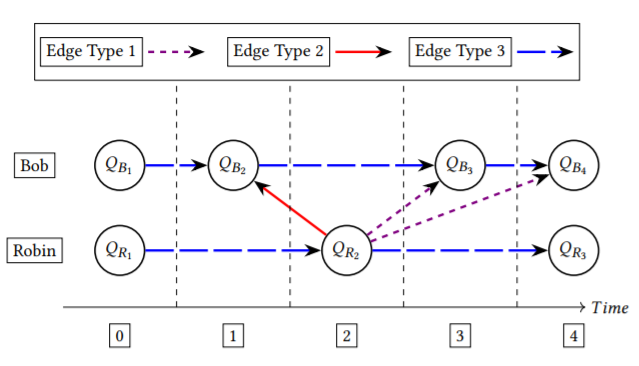}  
           \vspace{-5mm}
           \caption{A example depicting the network construction of \system. Here Bob has answered	 Robin's question $Q_{R_2}$.}
               \label{fig:example}
               \vspace{-5mm}
\end{figure}

\subsection{Parameters for Network Construction} There are two parameters to construct the network:

\noindent$\bullet$ {\bf Time of posting:} 
Each post (question/answer) is associated with an (absolute) posting time which is used to decide the edge type. Here, instead of considering the `absolute time' as the posting time, we divide the entire timespan present in a dataset into 2-week intervals (called `buckets'). We assume that those posts which were posted within a bucket have {\em same posting time}, and therefore, Hypotheses 1 and 2 may connect them in the network. The reasons behind choosing bucketed time instead of absolute time are two-fold: (i) we believe that questions posted by a user around the same time would be of same difficulty as s/he would not have gained much expertise on that topic in such a short period of time; (ii) considering absolute time may increase the number of edges in the network, which may unnecessarily densify the network and make the entire system memory intensive.  We chose bucketed interval as 2 weeks as it produces  the best result over a range from 1 week to 6 weeks (Table \ref{tab:parameter}). The bucket size matters for constructing the network as well as measuring few features (mentioned in Section \ref{model}).\\
$\bullet$ {\bf Recency of questions for type 2 edges:} For edge type 2, we have introduced $\delta_t$ to quantify the recency / contemporariness of Bob's question $Q_{B_i}$ (see the description of edge type 2). $Q_{B_i}$ has been asked `very recently' before Robin's question $Q_{R}$ which Bob has correctly answered.
We vary $\delta_t$ from $1$ to $5$ and observe that $\delta_t=1$ produces the best result (Table \ref{tab:parameter}).

\begin{table}
  \centering
    \begin{tabular}{c|c|c|c}
     \hline
          \multirow{2}{*}{{\bf Dataset}} & \multicolumn{3}{|c}{{\bf Edge count}} \\  \cline{2-4}
          & {\bf Edge type 1} & {\bf Edge type 2} & {\bf Edge type 3} \\ 
         \hline
         SO1 & 749,757 & 61,209 & 133,243 \\
         SO2 & 1,168,490 & 101,196  & 392,743 \\
         SO3 & 556,511 & 42,010  & 319,222  \\
         MSE & 224,058 & 10,124 & 89,996 \\
         \hline
  \end{tabular}
      \caption{Number of edges of each type in different datasets.}
      \label{edges}
      \vspace{-3mm}
      \end{table}

\begin{example}
Let us consider that Robin has asked three questions, $Q_{R_1}$, $Q_{R_2}$ and $Q_{R_3}$ at time $0$, $2$ and $4$ respectively, and Bob has correctly answered $Q_{R_2}$. Bob has also asked four questions $Q_{B_1}$, $Q_{B_2}$, $Q_{B_3}$ and $Q_{B_4}$ at time $0$, $1$, $3$ and $4$ respectively. Figure \ref{fig:example} shows the corresponding network for this example considering $\delta_t=1$.
\end{example}

\if{0}
\begin{table*}[!t]
\centering
\begin{tabular}{p{5.5cm}|p{5.5cm}|p{5.5cm}}
\hline
{\bf Hypothesis 1} & {\bf Hypothesis 2} & {\bf Hypothesis 3}\\\hline
\underline{Sample 1} & \underline{Sample 2} & \underline{Sample 3}\\
{\bf (Q1)} Write a program to generate all elements of power set? & {\bf (Q1)} Given an array of size n, give a deterministic algorithm (not quick sort) which uses O(1) space (not median of medians) and find the K\'th smallest item. & {\bf (Q1)} Given an array of integers (each <= $10^{6}$) what is the fastest way to find the sum of powers of prime factors of each integer?\\
{\bf (Q2)} How can one iterate over the elements of a stack starting from the top and going down without using any additional memory. The default iterator() goes from bottom to top. For Deque, there is a descendingIterator. Is there anything similar to this for a stack. If this is not possible which other Java data structures offer the functionality of a stack with the ability to iterate it backwards? 
& {\bf (Q2)} Transform a large file where each line is of the form: b d, where b and d are numbers. Change it from:
b -1 to b 1, where b should remain unchanged. There is a huge file and a way is required to achieve this using, say, sed or a similar tool?
& {\bf (Q2)} Given an array of N positive elements, one has to perform M operations on this array. In each operation a subarray (contiguous) of length W is to be selected and increased by 1. Each element of the array can be increased at most K times. One has to perform these operations such that the minimum element in the array is maximized. Only one scan is required.\\\hline
{\bf Response:} $diff$(Q2)>$diff$(Q1)? Yes: 75\%  & {\bf Response:} $diff$(Q2)>$diff$(Q1)? Yes: 75\% & {\bf Response:} $diff$(Q2)>$diff$(Q1)? Yes: 80\% \\\hline\hline
\underline{Random sample 1} & \underline{Random sample 2} & \underline{Random sample 3}\\ 
{\bf (Q1)} If an object implements the Map interface in Java and one wish to iterate over every pair contained within it, what is the most efficient way of going through the map? & {\bf (Q1)} Given a string s of length n, find the longest string t that occurs both forward and backward in s. e.g, s = yabcxqcbaz, then return t = abc or t = cba. Can this be done in linear time using suffix tree? & {\bf (Q1)} What happens at compile and runtime when concatenating an empty string in Java?
\\
{\bf (Q2)} How to format a number in Java? Should the number be formatted before rounding? 
& {\bf (Q2)} Given two sets A and B, what is the algorithm used to find their union, and what is it's running time?
& {\bf (Q2)} What is the simplest way to convert a Java string from all caps (words separated by underscores) to CamelCase (no word separators)?
\\\hline
{\bf Response:} $diff$(Q2)>$diff$(Q1)? Yes: 20\%  & {\bf Response:} $diff$(Q2)>$diff$(Q1)? Yes: 35\% & {\bf Response:} $diff$(Q2)>$diff$(Q1)? Yes: 30\% \\\hline\hline
\end{tabular}
\caption{One example question pair for each sample and its corresponding human response.} 
\label{surveyy}
\end{table*}

\fi

\subsection{Hypothesis Testing}\label{sec:hypothesis}
We further conducted a thorough  survey to show that three hypotheses behind our network construction mechanism are statistically significant.
For this, we prepared $6$ sets of edge samples, each two generated for testing each hypothesis as follows:
\begin{enumerate*}
    \item Sample 1: Choose $20$ edges of type 1 randomly from the network. 
    \item Random Sample 1: Randomly select $20$ pairs of questions (may not form any edge) which obey the time constraint mentioned in Hypothesis \ref{hyp1}.
    \item Sample 2: Choose $20$ edges of type 2 randomly from the network.
    \item Random Sample 2: Randomly select $20$ pairs of questions such that they follow the recency constraint mentioned in Hypothesis \ref{hyp2}.
    \item Sample 3: Choose $20$ edges of type 3 randomly from the network.
    \item Random Sample 3: Randomly select $20$ pairs of questions such that they follow the time constraint mentioned in Hypothesis \ref{hyp3}.
\end{enumerate*}

The survey was conducted with $20$ human annotators of age between 25-35, who are experts on Java and math-related domains. For a question pair ($Q_x,Q_y$) taken from each sample, we hypothesize that $diff(Q_y)>diff(Q_x)$. Null hypothesis rejects our hypothesis. Given a question pair, we asked each annotator to mark `Yes' if our hypothesis holds; otherwise mark `No'. One example question pair for each sample and the corresponding human response statistics (percentage of annotators accepted our hypothesis) are shown in Table \ref{surveyy}. Table \ref{pvalue} shows that for each hypothesis, the average number of annotators who accepted the hypothesis is higher for the sample taken from our network as compared to that for its corresponding random sample (the difference is also statistically significant as $p<0.01$, Kolmogorov-Smirnov test). This result therefore justifies our edge construction mechanism and indicates that all our hypotheses are realistic.

\begin{table*}[!h]
\centering
\scalebox{0.8}{
\begin{tabular}{p{5.5cm}|p{5.5cm}|p{5.5cm}}
\hline
{\bf Hypothesis 1} & {\bf Hypothesis 2} & {\bf Hypothesis 3}\\\hline
\underline{Sample 1} & \underline{Sample 2} & \underline{Sample 3}\\
{\bf (Q1)} Write a program to generate all elements of power set? & {\bf (Q1)} Given an array of size n, give a deterministic algorithm (not quick sort) which uses O(1) space (not median of medians) and find the K\'th smallest item. & {\bf (Q1)} Given an array of integers (each <= $10^{6}$) what is the fastest way to find the sum of powers of prime factors of each integer?\\
{\bf (Q2)} How can one iterate over the elements of a stack starting from the top and going down without using any additional memory. The default iterator() goes from bottom to top. For Deque, there is a descendingIterator. Is there anything similar to this for a stack. If this is not possible which other Java data structures offer the functionality of a stack with the ability to iterate it backwards? 
& {\bf (Q2)} Transform a large file where each line is of the form: b d, where b and d are numbers. Change it from:
b -1 to b 1, where b should remain unchanged. There is a huge file and a way is required to achieve this using, say, sed or a similar tool?
& {\bf (Q2)} Given an array of N positive elements, one has to perform M operations on this array. In each operation a subarray (contiguous) of length W is to be selected and increased by 1. Each element of the array can be increased at most K times. One has to perform these operations such that the minimum element in the array is maximized. Only one scan is required.\\\hline
{\bf Response:} $diff$(Q2)>$diff$(Q1)? Yes: 75\%  & {\bf Response:} $diff$(Q2)>$diff$(Q1)? Yes: 75\% & {\bf Response:} $diff$(Q2)>$diff$(Q1)? Yes: 80\% \\\hline\hline
\underline{Random sample 1} & \underline{Random sample 2} & \underline{Random sample 3}\\ 
{\bf (Q1)} If an object implements the Map interface in Java and one wish to iterate over every pair contained within it, what is the most efficient way of going through the map? & {\bf (Q1)} Given a string s of length n, find the longest string t that occurs both forward and backward in s. e.g, s = yabcxqcbaz, then return t = abc or t = cba. Can this be done in linear time using suffix tree? & {\bf (Q1)} What happens at compile and runtime when concatenating an empty string in Java?
\\
{\bf (Q2)} How to format a number in Java? Should the number be formatted before rounding? 
& {\bf (Q2)} Given two sets A and B, what is the algorithm used to find their union, and what is it's running time?
& {\bf (Q2)} What is the simplest way to convert a Java string from all caps (words separated by underscores) to CamelCase (no word separators)?
\\\hline
{\bf Response:} $diff$(Q2)>$diff$(Q1)? Yes: 20\%  & {\bf Response:} $diff$(Q2)>$diff$(Q1)? Yes: 35\% & {\bf Response:} $diff$(Q2)>$diff$(Q1)? Yes: 30\% \\\hline\hline
\end{tabular}}
\caption{One example question pair for each sample and its corresponding human response to evaluate the proposed hypotheses.} 
\label{surveyy}
\end{table*}

\begin{table}
  \centering
    \begin{tabular}{c|c|c|c}
     \hline
 {\bf Hypothesis} & {\bf Original sample} & {\bf Random sample} & {\bf $p$-value}  \\\hline
 H1 &  13.25 & 10.44 & $p$ \textless 0.01 \\

 H2 & 12.12 & 10.11 & $p$ \textless 0.01 \\
 
 H3  & 15.40  & 10.66 & $p$ \textless 0.01 \\  \hline
  \end{tabular}
     \newline
      \caption{Average number of annotators who accepted the hypotheses, and the $p$-value indicating the significance of our hypotheses w.r.t the null hypothesis.}
      \label{pvalue}
      \end{table}

\subsection{Edge Directionality Prediction Problem}\label{model}
Once the network is constructed, \system~considers the `relative question difficulty estimation' problem as an `edge directionality prediction' problem. Since an edge connecting two questions in a network points to the difficult question from the easy question, given a pair of questions with unknown difficulty level, the task boils down to predicting the direction of the virtual edge connecting these two questions in the network.

{\color{black} The mapping of the given problem  to the network-aided edge directionality prediction problem helps in capturing the overall dependency (relative difficulty) of questions. Edges derived based on the novel hypotheses lead to the generation of a directed network structure which further enables us to leverage the structural property of the network. This also helps in generating a feature vector, where each feature captures a different notion of difficulty of a question. Features allow to pool in the network topology, metadata and textual content in order to predict the relative difficulty of a question among a pair. The aim to map the given problem to the link prediction problem is to find a model which captures the role of each of the chosen features in predicting the relative hardness of question.}

Although research on link prediction is vast in network science domain \cite{LU20111150}, the problem of edge directionality prediction is relatively less studied. Guo et al.  \cite{Guolink-prediction} proposed a ranking-based method that combines both local indicators and global hierarchical structures of networks for predicting the direction of edge.  Soundarajan and Hopcroft \cite{Soundarajan} treated this problem as a supervised learning problem. They calculated various features of each known links based on its position in the network, and used SVM to predict the unknown directions of edges. Wang et al. \cite{Wang_link_prediction} proposed local directed path that solves the information loss problem in sparse network and makes the method effective and robust.


We also consider the `edge directionality prediction problem' as a supervised learning problem. The pairs of questions, each of which is directly connected via an edge in our network, form the training set.  
If the pair of questions under inspection is already connected directly by an edge, the problem will be immediately solved by looking at the directionality of the edge. In our supervised model, given a  pair of questions ($a,b$) (one entity in the population), we first connect them by an edge if they are not connected and then determine the directionality of the edge. Our supervised model uses the following features which are broadly divided into three categories: (i) {\em network topology based} (F1-F6), (ii) {\em metadata based} (F7-F10), and (iii) {\em textual content based} (F11-F12) 

\begin{itemize}
    \item {\bf [F1] Leader Follower ranking for node $a$:}  {\color{black} Guo et al. \cite{Guolink-prediction} used a ranking scheme to predict the missing links in a network. The pseudo-code is presented in Algorithm \ref{algo:leader_follower}. The algorithm takes an adjacency matrix adjMatrix(i)(j), $\forall i, j \in Nodes$ which is $1$ if there is edge between $i$ to $j$ and $0$ otherwise. The indegree  and outdegree of a node $i$ are represented by $in\_degree(i)$ and $out\_degree(j)$ respectively. At each step, for each node, $\gamma$, the difference between its in-degree and out-degree,  is computed to separate leaders (high-ranked nodes) from followers (low-ranked nodes). Apart from adjacency matrix, the algorithm expects $\alpha$ as input (set $0.65$ in our case as suggested in \cite{Guolink-prediction}), specifying the proportion of leaders at each step. The whole network is partitioned into leaders and followers, and this process continues recursively on the further groups obtained. If at any time, the number of nodes in any of the groups, $\|V\|$ is less than $1/\alpha$, then ranking is done according to $\gamma$. During the merging of leaders and followers, followers are placed after leaders. Finally, any node ranked lower (less important) is predicted to have an edge to higher ranked nodes (more important). In our case, the ranking is normalized by the number of questions and then used as a feature.}

   {\color{blue}
   \SetKwInput{KwInput}{Input}
    \SetKwInput{KwOutput}{Output}
    \SetKwRepeat{Repeat}{repeat}{until}
  \begin{algorithm}

    \SetAlgoLined
    \KwInput{{\color{black} adjMatrix($i$)($j$), $\forall i, j \in Nodes$, $\alpha$}} 
    \KwOutput{{\color{black} R($i$), $\forall i \in Nodes$}}
     
    {\color{black}
       $in\_degree(i) = count(adjMatrix(j)(i) == 1)$, $\forall i, j \in Nodes$\\
       $out\_degree(i) = count(adjMatrix(i)(j) == 1)$ , $\forall i, j \in Nodes$\\     
       $\gamma(i) = in\_degree(i) - out\_degree(i), \forall i \in Nodes$. \\
      Sort the nodes by descending order of $\gamma(j)$.\\
      For any node $(j)$ at position $p$, $I_j = p$ $(1 \leq p \leq \|Nodes\|)$. \\
      If $I_i \leq \alpha \|Nodes\|$, then $i \in Leaders$, else $i \in Followers, \forall i \in Nodes$. \\
      $R(i) = LF\_Rank(Leaders_i), \forall i \in Leaders$ \\
      $R(j) = \|Leaders\| + LF\_Rank(Followers_j), \forall j \in Followers$
    } 
 
    \caption{{\color{black} Leader Follower Ranking}}\label{algo:leader_follower}
    \end{algorithm}
    }
    \item {\bf [F2] Leader Follower ranking for node $b$:} We use the similar strategy to measure the above rank for node $b$. 
    \item {\bf [F3] PageRank of node $a$:} It emphasizes the importance of a node, i.e., a probability distribution signifying if a random walker will arrive to that node.
We use PageRank to compute the score of each node. However, we modify the initialization of PageRank by incorporating the weight of nodes -- let $A$ be the user asking question $Q_A$ and the reputation (normalized to the maximum reputation of a user) of $A$ be $r_{A}$. Then the PageRank score of $Q_A$ is calculated as:
\begin{equation}
     PR(Q_A) ={(1-d)} {r_{A}} + d \sum_{Q_j \in N(Q_A)} \frac{PR (Q_j)}{Outdegree(Q_j)}
\end{equation}
Here, $N(Q_A)$ is the neighbors of $Q_A$ pointing to $Q_A$; the damping factor $d$ is set to $0.85$.

\item {\bf [F4] PageRank of node $b$:} We similarly calculate the PageRank score of node $b$ as mentioned in F3. 

    \item {\bf [F5] Degree of node $a$:} It is computed after considering the undirected version of the network, i.e., number of nodes adjacent to a node.
    \item {\bf [F6] Degree of node $b$:} We calculate the degree of node $b$ similarly as mentioned in F5.    
 
    \item {\bf [F7] Posting time difference between $a$ and its accepted answer:} It signifies the difference between $T_a$, the posting time of $a$  and $T_{a'}$, the posting time of its accepted answer $a'$, if any. If none of the answers is accepted, it is set to 1. However, instead of taking the direct time difference, we employ an exponential decay as:
    $1 - e^{-(T_{a'}-T_a)}$.   
The higher the difference between the posting time and the accepted answer time (implying that users have taken more times to answer the question), the  more the difficulty level of the question. 

    \item {\bf [F8] Posting time difference between $b$ and its accepted answer:} Similar score is computed for $b$ as mentioned in F7.

    \item {\bf [F9] Accepted answers of users who posted $a$ till $T_a$:} The more the number of accepted answers of the user asking question $a$ till $a$'s posting time $T_a$, the more the user is assumed to be an expert and the higher the difficulty level of $a$. We normalize this score by the maximum value among the users.
    \item {\bf [F10] Accepted answers of users who posted $b$ till $T_b$:} Similar score is measured for question $b$ as mentioned in F9.

    \item {\bf [F11] Textual feature of $a$:} This feature takes into account the text present in $a$. The idea is that if a question is easy, its corresponding answer should be present in a particular passage of a text book. Therefore, for $a$, we first consider its accepted answer and check if the unigrams present in that answer are also available in different books.  For Java-related questions presents in SO1, SO2 and SO3, we consider the following books: (i) Core Java Volume I by Cay S. Horstmann \cite{Gvero:2013}, (ii) Core Java Volume II by Cay S. Horstmann \cite{jbook2}, (iii) Java: The Complete Reference by Herbert Schildt \cite{jbook3}, (iv) OOP - Learn Object Oriented Thinking and Programming by Rudolf Pecinovsky \cite{jbook4}, and (v)  Object Oriented Programming using Java by Simon Kendal \cite{jbook5}. For Math-related questions present in MSE, we consider the following books: (i) Advanced Engineering Mathematics by Erwin Kreyszig \cite{mbook1}, (ii) Introduction to Probability and Statistics for Engineers and Scientists by Sheldon M. Ross \cite{mbook2}, (iii) Discrete Mathematics and Its Applications by Kenneth Rosen \cite{mbook3}, (iv) Higher Engineering mathematics by B.S. Grewal \cite{mbook4}, and (v) Advanced Engineering Mathematics  by K. A. Stroud \cite{mbook5}. For each type of questions (Java/Math), we merge its relevant books and create a single document. After several pre-processing (tokenization, stemming etc.) we divide the document into different paragraphs, each of which forming a passage. Then we measure TF-IDF based similarity measure between each passage and the accepted answer. Finally, we take the maximum similarity among all the passages. The intuition is that if an answer is straightforward, most of its tokens should be concentrated into one passage and the TF-IDF score would be higher as compared to the case where answer tokens are dispersed into multiple passages. If there is no accepted answer for a question, we consider this feature to be $0$.  
    
\item {\bf [F12] Textual feature of $b$:}
We apply the same technique mentioned in F11 and measure the textual feature of $b$.
\end{itemize}

{\color{black} The proposed features are easy to compute once the network is constructed. Few features such as posting time, textual content do not require the network structure and can be derived directly from the metadata information. These features are chosen after verifying their significance on the data. They capture different notions of difficulty on a temporal scale.  
Next, we describe how these features can be implemented taking `usability' and `scalability' into account. Features which require one time computation such as degree, posting time difference between question and its accepted answer, accepted answers of question at a particular time stamp can be stored in a database (such as DynamoDB). We can compute PageRank and the leader-follower ranking  in distributed way using mapreduce as shown in \cite{Malewicz:2010}. The specified components are extremely fast and cost-effective. Such engineering tricks further help us in developing a real system mentioned in Section \ref{sec:system}.} 

In our supervised model, for each directed edge ($a,b$) ($b$ is more difficult than $a$), we consider ($a,b$) as an entity in the positive class (class 1) and ($b,a$) as an entity in the negative class (class 2). Therefore, in the training set the size of class 1 and class 2 will be same and equal to the number of directed edges in the overall network. We use different type of classifiers, namely SVM, Decision Tree, Naive Bayes, K Nearest Neighbors and  Multilayer Perceptron; among them SVM turns out to be the best model (Table \ref{tab:parameter}(c)).

\textbf{Time Complexity:}
Let $n$ denote the number of questions, $k$ be a constant which depends on textual content and $m$ denote the number of edges.
Both the hypotheses, H1 and H2, take $\mathcal{O}(n^{2})$ times while H3 is computed in $\mathcal{O}(n)$. For the different features we used, leader-follower Ranking takes $\mathcal{O}(n\log{}n)$ time, PageRank takes $\mathcal{O}(n + m)$ time, degree takes $\mathcal{O}(n)$ time, Posting time difference between question and its accepted answer takes $\mathcal{O}(n)$ time, Accepted answers of users who posted question takes $\mathcal{O}(n)$ time and textual feature of Question takes $\mathcal{O}(nk)$ time. Thus, the overall time complexity turns out to be $\mathcal{O}(n^{2})$.

\vspace{-2mm}
\subsection{Transitive Relationship}
An important issue is the transitive relations among questions whose difficulties are determined in a pairwise manner. Let $Q_i$, $Q_j$, and $Q_k$ be three questions, and let \system~produce the relative difficulty as $diff(Q_k)>diff(Q_j)>diff(Q_i)$. Transitivity would then imply that $diff(Q_k)>diff(Q_i)$; whereas non-transitivity condition implies that  $diff(Q_i)>diff(Q_k)$ which in turn creates cycles in the underlying graph.

To test the presence of transitive vs non-transitive cases, we set out to find non-transitive cases first, i.e., cases where cycle (of different length) exists. We randomly pick a set of $n$ nodes 10,000 times, where $n$ indicates the size of the cycle we want to investigate. For each set, we check if the cycle exists in any permutation of chosen nodes. We repeat the experiment by varying $n$ from $2$ to $10$. We observed that for any value of $n$, in less than $1\%$ of cases, the cycle exists and in more than $99\%$ cases transitivity holds. Our claim turns out to be statistically significant with $p<0.01$.

\section{Experimental Results}\label{sec:result}
This section presents the performance of the competing methods. It starts by briefly describing the human annotation for test set generation, followed by the baseline methods. It then thoroughly elaborates the parameter selection process for \system, comparative evaluation of the competing methods and other important properties of \system.  All computations were carried out on single Ubuntu machine with $32$ GB RAM and $2.7$GHz CPU with $12$ cores.

\subsection{Test Set Generation}
{\color{black} For each dataset mentioned in Section \ref{sec:dataset}, we prepared a test set for evaluating the competing models as follows.
Three experts\footnote{All of them also helped us validating our hypotheses  as discussed in Section \ref{sec:hypothesis}.} were given 1000 question pairs to start the annotation process. While annotating, experts were shown pairs of questions along with their description. We had taken complete care of keeping metadata information (such as the information of the questioners, upvote count of answers, etc.) secret. All the three experts were told to separately mark each pair of questions ($q_1$, $q_2$) in terms of their relative difficulty: $q_1$ is tougher than $q_2$, $q_1$ is easier than $q_2$ or difficult to predict. The moderator stopped the annotation process once $300$ pairs were annotated for each data such that all three annotators agreed upon their annotations (i.e., the complete consensus was achieved for each pair). The annotators ended up annotating $356$, $350$, $362$ and $358$ pairs to reach a consensus on $300$ pairs for SO1, SO2, SO3 and MSE respectively.  The inter-annotator agreement using Fleiss' kappa is $0.84$. We used total $1,200$ such annotated pairs from four datasets for comparative evaluation.}

\subsection{Baseline Models and Evaluation Measures}
We consider four baselines described below; first three are existing methods (see Section \ref{sec:rw:qd})  and the last one is designed by us:
\begin{itemize}

\item {\bf Trueskill:} The approach was proposed by Liu et al. \cite{msr-paper}. It creates a user network which takes into account elements such as the question (pseudo user), question asker, best answerer user and each of the non-best answerers users. For each question, it creates four competitions/pairs: one between pseudo user U$_{Q}$ and asker U$_{a}$, one between pseudo user U$_{Q}$ and the best answerer U$_{b}$, one between the best answerer U$_{b}$ and asker U$_{a}$ and one between the best answerer U$_{b}$ and each of the non-best answerers of the question Q. This creates a user model, and all the competitions are fed to TrueSkill, a bayesian Skill rating model which calculates 
rank of each node in the model. 
For a given pair of questions, the question with higher rank (generated by Trueskill)  is considered difficult than the question with lower rank. 

\item {\bf RCM:} The Regularized Competition Model (RCM) was proposed by \citet{regularised}. It uses the network construction by \cite{msr-paper}. It solves the major problem of data sparseness i.e., low count of edges between questions present in \cite{msr-paper}. It assigns difficulty to every question and user. To overcome the sparsity problem, it further enhances knowledge by using textual similarity of question's description (title, body, tags) in inferring appropriateness of difficulty assigned. So, a pair of textually similar question should be of similar difficulty. It then proposes a regularized competition model  to assign a difficulty score. The algorithm tries to minimize loss on two fronts. First, if there is an edge in network between two nodes, then their should be at least $\delta$ difference between its difficulties. Second, if two questions are similar as per their textual descriptions, then the difference in their difficulties should be minimum. It forms a vector $\theta \in \mathcal{R}^{M+N}$ where initial $M$ entries are expertise of users and next $N$ entries are difficulty of questions. A competition set $C$ is defined as a set consisting of all edges in the network. The algorithm essentially tries to minimize: $\sum_{C} l(\theta_i, \theta_j)$ where (i,j) denotes an edge in network and $ l(\theta_i, \theta_j) = max(0, \delta - (\theta_j - \theta_i))$. It also tries to minimize $\sum_{i=1}^{N} \sum_{j=1}^{N} w_{\text{ij}} (\theta_j - \theta_i)^2  $ for every question pair (i,j) where $w_{\text{ij}}$ indicates similarity between question description. The algorithm starts at initial $\theta$ and proceeds towards negative subgradient, $\theta_{t+1}$ = $\theta_t$ - $\gamma_t$*$\triangledown$$\mathcal{L}$($\theta_t$), where $\triangledown$$\mathcal{L}$($\theta_t$) is the subgradient and $\gamma_t$ is set as 0.001.
\item {\bf PageRank:} The approach was proposed by \citet{chinese-paper}. It exploits the concept of acceptance of answer in CQA Services in constructing the model.
It constructs the model in a way such that there is a directed edge from the question(s) where a particular user's answer is accepted to each question where his/her answer is not accepted.
Suppose user B answers questions Q1, Q2 and Q3 wherein answer to Q1 is only accepted. Then there will be a directed edge from Q2 to Q1, and Q3 to Q1. PageRank is then applied on the generated graph to get a score for each node (question).
For a given pair of questions, a question with higher PageRank score indicates a higher difficulty level than the question with lower PageRank score. 

%
\item {\bf HITS:} We further propose a new baseline as follows -- we run HITS algorithm \cite{hits} on our network and rank the questions globally based on their authoritativeness. We use authority score instead of hub score to evaluate the difficulty of question based on the node itself. The importance of links is captured in construction of network. Now given a pair of questions, we mark the one as more difficult whose authoritativeness is higher. 

\end{itemize}

We measure the accuracy of the methods based on F\textsubscript{1} score and Area under the ROC curve (AUC).

\subsection{Parameter Selection for \system}
There are three important parameters of \system: (i) bucket size for determining question posting time, (ii) recency of questions for type 2 edges ($\delta_t$) and (iii) classifier for edge directionality prediction. Table \ref{tab:parameter} shows   F\textsubscript{1} score of \system~by varying the parameter values on SO3 dataset\footnote{The pattern was same for the other datasets and therefore not reported here.}. We vary the bucket size from $1$ to $6$ and observe that $2$-week bucket interval produces the best result. {\color{black}$\delta_t$ is varied from $1$ to $8$ and $\delta_t=1$ gives the highest accuracy.} We use seven classifiers: SVM, Decision Tree (DT), Naive Bayes (NB), K-Nearest Neighbors (KNN), Multilayer Perceptron (MLP), Random Forest(RF), and Gradient Boosting (GB) with suitable hyper-parameter optimization; among them SVM (with RBF kernel) produces the best accuracy.  Therefore, unless otherwise stated, {\bf we use the following parameters as default for \system: bucket size=2, $\delta_t=1$ and SVM classifier}.

\begin{table}
\centering
\begin{tabular}{c|c|c|c|c|c}
\multicolumn{6}{c}{{\bf (a) Bucket size}}\\\hline
1 & 2 & 3 & 4 & 5 & 6\\\hline
70.1 & {\bf 76.39} & 71.1 & 68.3 & 70.1 & 60.4 \\\hline
\end{tabular}

\begin{tabular}{c|c|c|c|c|c|c|c}
\multicolumn{8}{c}{{\bf (b) $\delta_t$}}\\\hline
1 & 2 & 3 & 4 & 5 & 6 & {\color{black}7} & {\color{black}8}\\\hline
{\bf 76.39} & 76.11 & 75.88 & 75.88 & 75.91 & 76.2 & {\color{black} 75.64} & {\color{black} 75.56}\\\hline
\end{tabular}

\begin{tabular}{c|c|c|c|c|c|c}
\multicolumn{7}{c}{{\bf (c) Classifier}}\\\hline
SVM & DT & NB & KNN & MLP & RF & GB\\\hline
{\bf 76.39}  & 63.9  & 75.2 & 69.1  & 75.4 & 72.0 & 75.0\\\hline
\end{tabular} 
\caption{F\textsubscript{1} score of \system~ with different parameter combinations on SO3 dataset. For each parameter, we vary its value keeping the other parameters default.}\label{tab:parameter}
\vspace{-2mm}
\end{table}

\begin{table*}[t!]
\centering
\resizebox{\columnwidth}{!}{
\begin{tabular}{|c|c|c|c|c|c|c||c|c|c|c|c|}
\hline
   \multirow{2}{*}{{\bf Dataset}} & {\bf Underlying} & \multicolumn{5}{|c||}{{\bf F\textsubscript{1} score (\%)}} &  \multicolumn{5}{|c|}{{\bf AUC (\%)}} \\ \cline{3-12}
          & {\bf Network} & {\bf RCM} & {\bf Trueskill} & {\bf PageRank} & {\bf HITS} & {\bf \system} & {\bf RCM} & {\bf Trueskill} & {\bf PageRank} & {\bf HITS} & {\bf \system}  \\
\hline
 
\multirow{3}{*}{SO1} & RCM & {\bf {\color{red}55.6}} & 54.0 & 62.8& 52.8&33.3 &{\bf {\color{red}55.7}} & 54.1& 62.8& 52.9&48.1 \\ 
& Trueskill &55.6 & {\bf 54.0}  & 62.8 & 52.8 & 33.3 &55.7 & {\bf 54.1} & 62.8 & 52.9 & 48.1\\ 
& PageRank &42.2 & 56.1 & {\bf 50.3} & 53.5 & 38.1 &42.3 & 56.2 & {\bf 50.5} & 53.5 & 50.5 \\ 
& \system & 56.5& 66.6 & 62.1 & {\bf 49.4}  & {\color{blue}{\bf 71.6}} &56.6 & 66.9 & 62.4 & {\bf 49.4} & {\bf {\color{blue}71.7}} \\\hline

\multirow{3}{*}{SO2} & RCM & {\bf {\color{red}57.6}} &52.8 &51.5 &49.2 &32.4 &{\bf  {\color{red}57.3}}& 52.9&51.6 &49.3 &50.0 \\
& Trueskill &57.6 & {\bf 52.8}  & 51.5 & 49.2 & 32.4 &57.6 & {\bf 52.9} & 51.6 & 49.3 & 50.1\\ 
& PageRank & 46.9& 49.3&{\bf 49.3} & 49.6& 34.2 &47.0 &49.4 &{\bf 49.4} &49.6 & 48.7 \\ 
& \system & 56.9 & 56.1 & 57.2 & {\bf 54.6} & {\color{blue}{\bf 70.5 }} & 56.9& 56.2 & 57.3 &{\bf 54.6} & {\bf {\color{blue}70.5}}\\ \hline

\multirow{3}{*}{SO3} & RCM & {\bf 50.7} &52.2 &54.6 &48.8 &36.9 & {\bf 50.7}&53.2 &55.4 &50.1 &50.0 \\ 
& Trueskill & 50.7& {\bf {\color{red}52.2}} & 54.6 & 48.8 & 36.9 &50.7 & {\bf {\color{red}53.2}} & 55.4 & 50.1 & 50.0\\ 
& PageRank & 51.7& 51.1 & {\bf 50.3} & 44.9 & 36.9 & 52.2& 51.4 & {\bf 51.0} & 46.0 & 50.0\\ 
& \system &54.1 & 60.7 &58.9 &{\bf 49.3} &  {\color{blue}{\bf 76.3}} &54.7 & 62.2 & 60.7& {\bf 50.8} & {\bf {\color{blue}77.1}}\\ \hline

\multirow{3}{*}{MSE} & RCM & {\bf 55.3} &58.6 &55.9 &52.6 &32.1 &{\bf 55.3} &58.6 &55.9 &52.6 &48.7 \\ 
& Trueskill & 55.3& {\bf 58.6} & 55.9 & 52.6 & 32.2 &55.3 & {\bf 58.6} & 55.9 & 52.6 & 48.7\\ 
& PageRank &49.3 & 51.7 & {\bf 53.6} & 56.2 & 34.6 & 49.4& 51.8 & {\bf 53.6} & 56.2 & 49.1 \\ 
& \system &55.6 & 54.8 &61.6 & {\bf {\color{red}58.9}} & {\color{blue}{\bf 71.8 }} & 55.7&54.8&61.6&{\bf{\color{red} 59.0}}& {\bf {\color{blue}72.3}}\\\hline 

\end{tabular}}
\newline
\caption{(Color online) Accuracy (in terms of F\textsubscript{1} and AUC) of the competing methods on four different datasets. \system~is run with its default configuration. Boldface numbers are the accuracy of the baseline methods using the configuration reported in the original papers. Blue (red) numbers are the accuracies of the best performing (second-ranked) method. We also measure the accuracy of each competing method using the network suggested by other methods and notice that most of the methods perform better if \system's network is fed into their models (which indeed shows the superiority of our network construction mechanism).}
\label{tab:result}
\vspace{-6mm}
\end{table*}

\subsection{Comparative Evaluation}
Table \ref{tab:result} shows the accuracy of all the competing methods across different datasets (see the boldface values only). Irrespective of the dataset and the evaluation measures, \system~outperforms all the baselines by a significant margin. \system~achieves a consistent performance of more than 70\% F\textsubscript{1} score and 70\% AUC across different datasets. However, none of the single baseline stands as the best baseline consistently across all the datasets (see the red numbers in Table \ref{tab:result}). \system~beats the best baseline by 28.77\% (28.72\%), 22.79\% (23.03\%), 46.16\% (44.92\%) and 21.90\% (22.54\%) higher F\textsubscript{1} score (AUC) for SO1, SO2, SO3 and MSE datasets respectively.  {\color{black} We also observe that only in SO3, Trueskill beats RCM. Trueskill is a Bayesian skill rating model, where skill level of pseudo-user (based upon the network construction) is modeled by a normal distribution and updated according to Bayes theorem. SO3, the Stack Overflow data corresponding to the most recent time frame, consists of most amount of data, in terms of both questions and users. Therefore, normal distribution may be able to capture the skill level of pseudo-user better in comparison to the optimization problem of RCM.}

Since all the baselines first create a network and run their difficulty estimation modules on that network, one may argue how efficient our network construction mechanism is as compared to other mechanisms. To check this, we further consider each such network (obtained from each competing method) separately, run the difficulty estimation module suggested by each competing method and report the accuracy in Table \ref{tab:result}. For instance, the value reported in third row and first column (i.e., $42.2$) indicates the accuracy of RCM on the network constructed by PageRank. We once again notice that in most cases, the competing methods perform the best with \system's network. This indeed shows the superiority of our network construction mechanism.  

{\color{black}
We also compare \system~with two learning-to-rank methods -- RankBoost \cite{freund2003efficient} and RankSVM \cite{lee2014large,joachims2002optimizing,herbrich1999support}, and a modified versions of PageRank and HITS (we call them Modified PageRank \cite{kumar2011page} and Modified HITS \cite{liu2012improved} respectively). The performance of these methods in terms of F\textsubscript{1} and AUC is reported in Table \ref{tab:newbaseline}. Once again, we observe that \system~significantly outperforms all these baselines across all the datasets.}

\begin{table}[!t]
    \centering
    \scalebox{0.8}{
    \begin{tabular}{|l|c|c|c|c|c|}
    \hline
         \multicolumn{1}{|c|}{{\bf Method}} & {\bf Metric} & {\bf SO1} & {\bf SO2} & {\bf SO3} & {\bf MSE}  \\\hline
         \multirow{2}{*}{RankBoost} & F\textsubscript{1} score& 64.21 & 64.24 & 63.44 & 63.24\\
         & AUC & 63.64 & 63.84 & 62.22 & 61.48\\\hline

         \multirow{2}{*}{RankSVM} & F\textsubscript{1} score & 61.62 & 62.46 & 61.82 & 61.21\\
         & AUC & 60.54 & 61.28 & 60.48 & 60.64\\\hline

         \multirow{2}{*}{Modified PageRank} & F\textsubscript{1} score & 62.53 & 58.25 & 58.90 & 60.52\\
         & AUC & 62.90 & 58.36 & 60.48 & 60.52\\\hline
         
     \multirow{2}{*}{Modified HITS} &  F\textsubscript{1} score & 61.46 & 60.54 & 59.81 & 61.21\\
     & AUC & 60.44 & 60.21 & 58.46 & 60.42\\\hline
     
     \multirow{2}{*}{\system} &  F\textsubscript{1} score & 71.60 & 70.56 & 76.39 & 71.80 \\
     & AUC & 71.70 & 70.50 & 77.10 & 72.30\\\hline

    \end{tabular}}
    \caption{Performance comparison of other baselines with \system.}
    \label{tab:newbaseline}
\end{table}

\subsection{Feature and Hypothesis Importance}
We also measure the importance of each feature  for \system~with default configuration. For this, we drop each feature in isolation and measure the accuracy of \system. Figure \ref{fig:factor_imp}(a) shows that the maximum decrease in accuracy (27.63\% decrease in F\textsubscript{1}) is observed when we drop leader follower ranking (F1 and F2), followed by PageRank (F3 and F4) and degree (F5 and F6). However, there is no increase in accuracy if we drop any feature, indicating that all features should be considered for this task.

One would further argue on the importance of keeping all hypotheses (all edge types in the network). For this, we conduct similar type of experiment -- we drop each edge type (by ignoring its corresponding hypothesis), reconstruct the network and measure the accuracy of \system. Figure \ref{fig:factor_imp}(b) shows  similar trend -- dropping of each edge type decreases the accuracy, which indicates that all types of edges are important. However, edge type 2  (Hypothesis 2) seems to be more important (dropping of which reduces 26.31\% F\textsubscript{1} score), followed by type 1 and type 3 edges.

\begin{figure}[t!]
    \centering
           \includegraphics[width=\columnwidth]{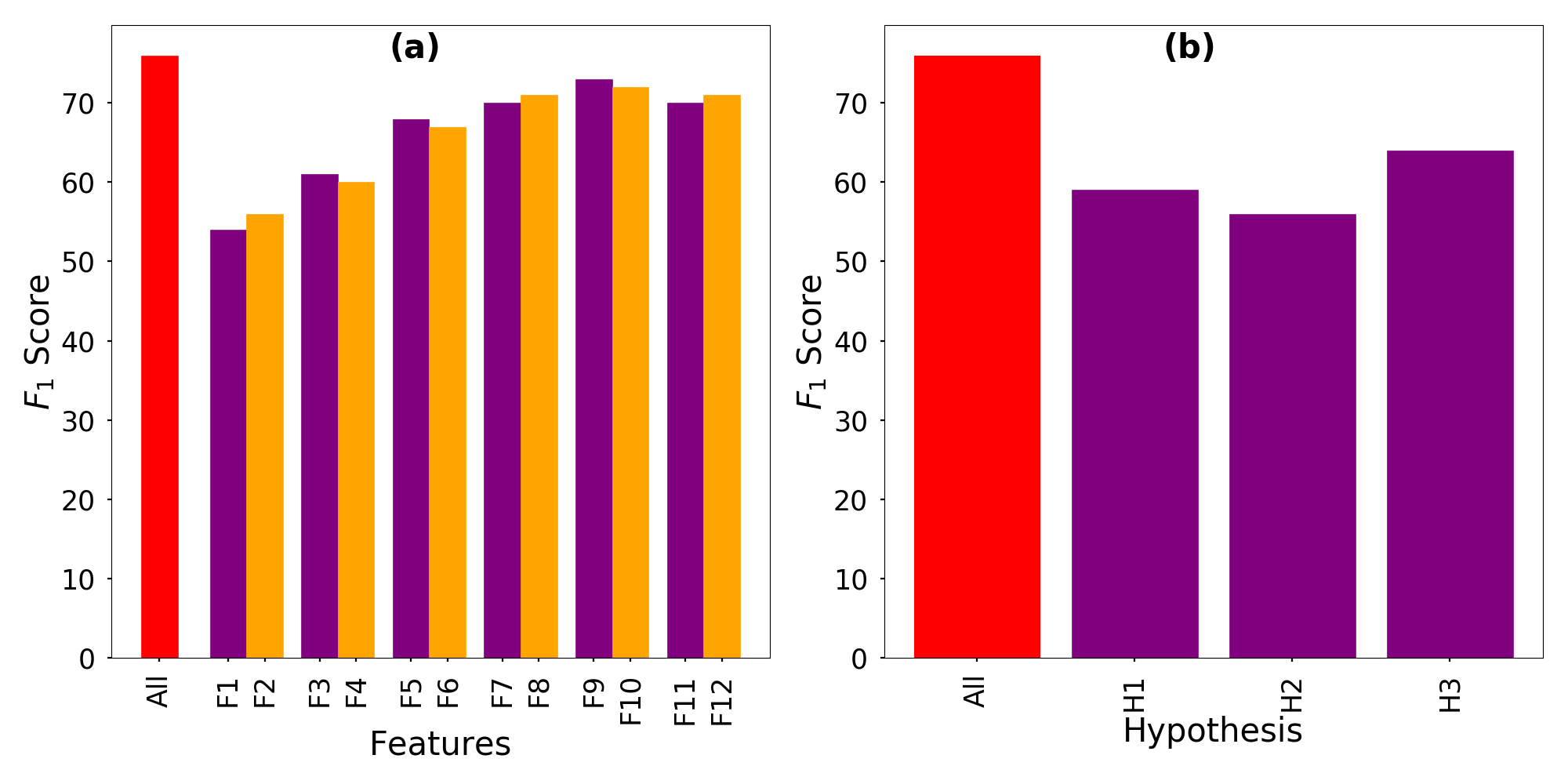}
           \caption{Importance of (a) features and (b) hypotheses based on F\textsubscript{1} score for SO3 dataset. For better comparison, we also show the results with all features and all hypotheses (All).
              \label{fig:factor_imp}}
\end{figure}

\subsection{Robustness under Noise}
\begin{figure}[!h]

    \centering
           \includegraphics[width=0.8\columnwidth]{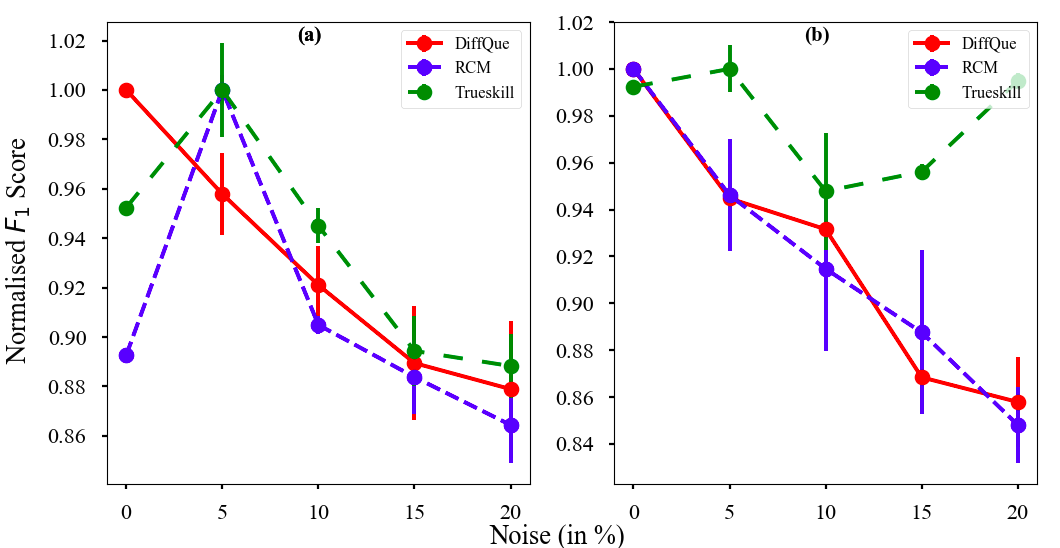}
           \caption{Change in the accuracy (F\textsubscript{1} score normalized by the maximum) of \system~, RCM  and Trueskill with the increase of (a) noise 1 and (b) noise 2  on the SO3 dataset.\\}
               \label{fig:noise3}
               
\end{figure}

We further test the robustness of \system~by employing two types of noises into the network: (i) {\bf Noise 1}, where we keep inserting $x\%$ of the existing edges randomly into the network, thus increasing the size of the training set, and (ii) {\bf Noise 2}, where we first remove $x\%$ of the existing edges and randomly insert $x\%$ of the edges into the network, thus the size of the training set remaining same after injecting the noise. We vary $x$ from $5$ to $20$ (with the increment of $5$). We hypothesize that a robust method should be able to tolerate the effect of noise up to a certain level, after which its performance should start deteriorating. 
Figure \ref{fig:noise3} shows the effect of both types of noises on the performance of \system, RCM (the best baseline), and Trueskill. \system~seems to perform almost the same until 5\% of both types of noises injected into the training set. However, the behavior of RCM is quite abnormal -- sometime its performance improves with the increase of noise level (Figure \ref{fig:noise3}(a)). Trueskill also follows the same pattern as RCM. Nevertheless, we agree that \system~is not appropriately resilient to noise as it does not seem to tolerate even less amount of noise (which opens further scope of improvement); however its performance does not seem to be abnormal.

\subsection{Capability of Domain Adaptation}
Another signature of a robust system is that it should be capable of adopting different domains, i.e., it should be able to get trained on one dataset and perform on another dataset. To verify this signature, we train \system~ and RCM (the best baseline) on one dataset and test them on another dataset. We only consider those users which are common in both training and test sets. For SO datasets, it is fairly straightforward; however when we choose cross-domain data for training and testing (MSE as training and SO as testing and vice versa), we use the `account id' of users as a key to link common user\footnote{SO and MSE come under Stack exchange (SE) (\url{https://stackexchange.com/}). Therefore, each user has to maintain a common `account id' to participate in all CQA services under SE.}. Table \ref{tab:domain} shows that the performance of \system~almost remains  the same even when the domains of training and test sets are different. 

{\color{black} Interestingly, we observe that \system\ performs better while SO1 and SO2 are used for training and SO3 for testing, in comparison to both training and testing on SO3. The reason may be as follows. In SO3, the number of type 1 (related to hypothesis 1) and type 2 (related to hypothesis 2) edges is less than that in SO1 and SO2. Note that as shown in Figure \ref{fig:factor_imp}(b) that hypotheses 1 and 2 are the most important hypotheses. Therefore, the lack of two crucial edge types in SO3 may not allow the model to get trained properly. This may be the reason that our model does not perform better while trained on SO3.  }

Table \ref{tab:domain1} further shows a comparison of \system~ with RCM  which also claims to support domain adaptation \cite{msr-paper}.  
As the network topology and metadata have significant impact on question difficulty, we observe that such factors are domain independent, and our algorithm captures them well and thus can be adapted across domains. Tables \ref{tab:domain2}, \ref{tab:domain3}, \ref{tab:domain4} show the capability of domain adaptation of three baselines -- Trueskill, Pagerank and HITS respectively.

\subsection{Handling Cold Start Problem}
{\color{black}Most of the previous research \cite{chinese-paper,msr-paper} deal with well-resolved questions which have received enough attention in terms of the number of answers, upvotes etc. However, they suffer from the `cold start' problem -- they are inefficient to handle newly posted questions with no answer and/or the users who posted the question are new to the community.  In many real-world applications such as question routing and incentive mechanism design, however, it is usually required that the difficulty level of a question is known instantly after it is posted \cite{regularised}.

\begin{table}[!t]
\centering
\scalebox{0.85}{
\begin{tabular}{|c|c|c|c|c|c|}
\hline
\multicolumn{6}{|c|}{{\bf Testing}}\\\hline
& & {\bf SO1} & {\bf SO2} & {\bf SO3} & {\bf MSE} \\\hline
\parbox[t]{2mm}{\multirow{4}{*}{\rotatebox[origin=c]{90}{{\bf Training}}}}& {\bf SO1}  & 71.65 & 66.99 & 76.68 & 71.49 \\\cline{2-6}
& {\bf SO2}  & 70.21 & 70.53 & 77.24 & 67.60 \\\cline{2-6}
&{\bf SO3}  & 70.27 & 66.65 & 76.39 & 71.22 \\\cline{2-6}
&{\bf MSE}  & 70.98 & 66.99 & 76.36 & 71.84 \\\hline
\end{tabular}}
\caption{F\textsubscript{1} score of \system~ for different combination of training and test sets.}\label{tab:domain}
\vspace{-3mm}
\end{table}

\begin{table}[!h]
\centering
\scalebox{0.85}{
\begin{tabular}{|c|c|c|c|c|c|}
\hline
\multicolumn{6}{|c|}{{\bf Testing}}\\\hline
& & {\bf SO1} & {\bf SO2} & {\bf SO3} & {\bf MSE} \\\hline
\parbox[t]{2mm}{\multirow{4}{*}{\rotatebox[origin=c]{90}{{\bf Training}}}}&
{\bf SO1}              & 55.60  & 55.60 & 43.98 & 48.88 \\\cline{2-6}
&{\bf SO2}              & 60.09 & 57.60  & 54.60 & 54.15 \\\cline{2-6}
&{\bf SO3}              & 56.15 & 50.31 & 50.70  & 53.35 \\\cline{2-6}
&{\bf MSE}              & 51.60 & 49.99 & 53.28 & 55.30\\\hline
\end{tabular}} 
\caption{F\textsubscript{1} score of RCM  for different combination of training and testing sets.}\label{tab:domain1}
\vspace{-5mm}
\end{table}

\begin{table}[!h]
\centering
\scalebox{0.85}{
\begin{tabular}{|c|c|c|c|c|c|}
\hline
\multicolumn{6}{|c|}{{\bf Testing}}\\\hline
& & {\bf SO1} & {\bf SO2} & {\bf SO3} & {\bf MSE} \\\hline
\parbox[t]{2mm}{\multirow{4}{*}{\rotatebox[origin=c]{90}{{\bf Training}}}}&
{\bf SO1}              & 54.00    & 52.3  & 52.35 & 58.97 \\\cline{2-6}
&{\bf SO2}              & 52.48 & 52.8  & 52.27 & 59.32 \\\cline{2-6}
&{\bf SO3}              & 52.3  & 52.49 & 52.2  & 57.65 \\\cline{2-6}
&{\bf MSE}              & 51.64 & 52.49 & 53.02& 58.6\\\hline
\end{tabular}}
\caption{F\textsubscript{1} score of Trueskill  for different combination of training and testing sets.}\label{tab:domain2}
\vspace{-5mm}
\end{table}

\begin{table}[!h]
\centering
\scalebox{0.85}{
\begin{tabular}{|c|c|c|c|c|c|}
\hline
\multicolumn{6}{|c|}{{\bf Testing}}\\\hline
& & {\bf SO1} & {\bf SO2} & {\bf SO3} & {\bf MSE} \\\hline
\parbox[t]{2mm}{\multirow{4}{*}{\rotatebox[origin=c]{90}{{\bf Training}}}}&
{\bf SO1}              & 50.30  & 49.01 & 43.98 & 52.44 \\\cline{2-6}
&{\bf SO2}              & 49.82 & 49.30  & 46.75 & 53.05 \\\cline{2-6}
&{\bf SO3}              & 49.51 & 49.41 & 50.30  & 53.77 \\\cline{2-6}
&{\bf MSE}              & 52.18 & 48.22 & 48.04 &53.6\\\hline
\end{tabular}} 
\caption{F\textsubscript{1} score of PageRank  for different combination of training and testing sets.}\label{tab:domain3}
\vspace{-5mm}
\end{table}

\begin{table}[!h]
\centering
\scalebox{0.85}{
\begin{tabular}{|c|c|c|c|c|c|}
\hline
\multicolumn{6}{|c|}{{\bf Testing}}\\\hline
& & {\bf SO1} & {\bf SO2} & {\bf SO3} & {\bf MSE} \\\hline
\parbox[t]{2mm}{\multirow{4}{*}{\rotatebox[origin=c]{90}{{\bf Training}}}}&
{\bf SO1}              & 49.40  & 54.63 & 49.39 & 57.84 \\\cline{2-6}
&{\bf SO2}              & 48.10  & 54.60  & 48.97 & 61.47 \\\cline{2-6}
&{\bf SO3}              & 52.54 & 51.34 & 49.30  & 58.42 \\\cline{2-6}
&{\bf MSE}              & 48.92 & 53.58 & 45.53 & 58.9\\\hline
\end{tabular}}
\caption{F\textsubscript{1} score of HITS  for different combination of training and testing sets. \\}\label{tab:domain4}
\end{table}

RCM model leverages the textual content to handle cold-start problem. It uses a boolean term weighting to represent each question using a feature vector. To determine the similar questions, RCM measures Jaccard coefficient between feature vectors and selects top K nearest neighbors. \system~also handles the cold start problem by exploiting the textual description of questions. In our network construction, a question which has been asked by a completely new user forms an isolated node; therefore none of the network related features (such as F1-F6) can be computed for that question. Moreover, if it does not receive any accepted answer, other features (F7-F12) cannot be measured. We call those questions ``brand-new'' for which none of the features can be computed, leading to extreme level of cold start. Given a question pair  ($Q_1, Q_2$) under inspection,  there can be two types of cold start scenarios: (i) both $Q_1$ and $Q_2$ are brand new, (ii) either $Q_1$ or $Q_2$ is brand-new. To handle the former, we run  Doc2Vec \cite{Le:2014}, a standard embedding technique on textual description of all the questions and return, for each of $Q_1$ and $Q_2$, $k$ most similar questions (based on cosine similarity)  which are not brand-new. For $Q_1$ ({\em resp.} $Q_2$), the $k$ most similar questions are denoted as $\{Q^i_1\}_{i=1}^k$ ({\em resp.} $\{Q^i_2\}_{j=1}^k$). We assume each such set as the representative of the corresponding brand-new question. We then run \system~to measure the relative difficulty of each pair (one taken from $\{Q^i_1\}$ and other take from $\{Q^i_2\}$). Finally, we consider that brand-new question to be difficult whose corresponding neighboring set is marked as difficult maximum times by \system. The latter cold start scenario is handle in a similar way. Let us assume that $Q_1$ is old and $Q_2$ is brand-new. For $Q_2$, we first extract $k$ most similar questions $\{Q^i_2\}$ using the method discussed above. We then run \system~ to measure the relative difficulty between $Q_1$ and each of $\{Q^i_2\}$, and mark the one as difficult which wins the most.          
 
 \begin{figure}[!t]
    \centering
           \includegraphics[width=\columnwidth]{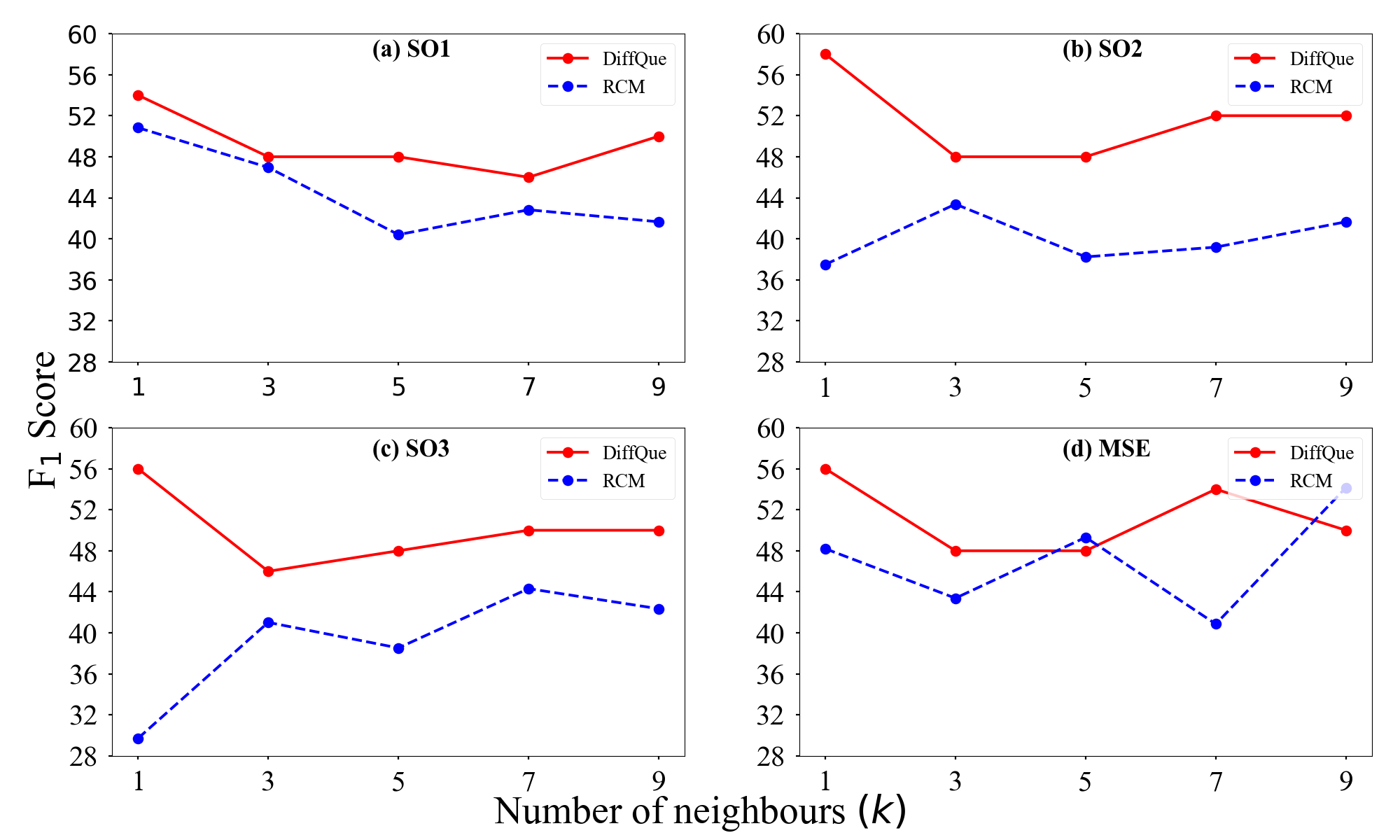}
           \caption{Accuracy of \system~and RCM  with different number of neighbors.}
               \label{fig:coldstart}
\end{figure}
 
To test the efficiency of our cold start module, we remove $50$ annotated pairs (edges and their associated nodes) randomly from the network. These pairs form the test set for the cold start problem.  Figure \ref{fig:coldstart} shows that  with the increase of $k$, \system~always performs better than RCM (the only baseline which handles cold start problem \cite{regularised}), indicating \system's superiority in tackling cold start problem.}

\section{System Description}\label{sec:system}
\begin{figure}[!t]
\centering
\fbox{
\includegraphics[width=0.7\columnwidth]{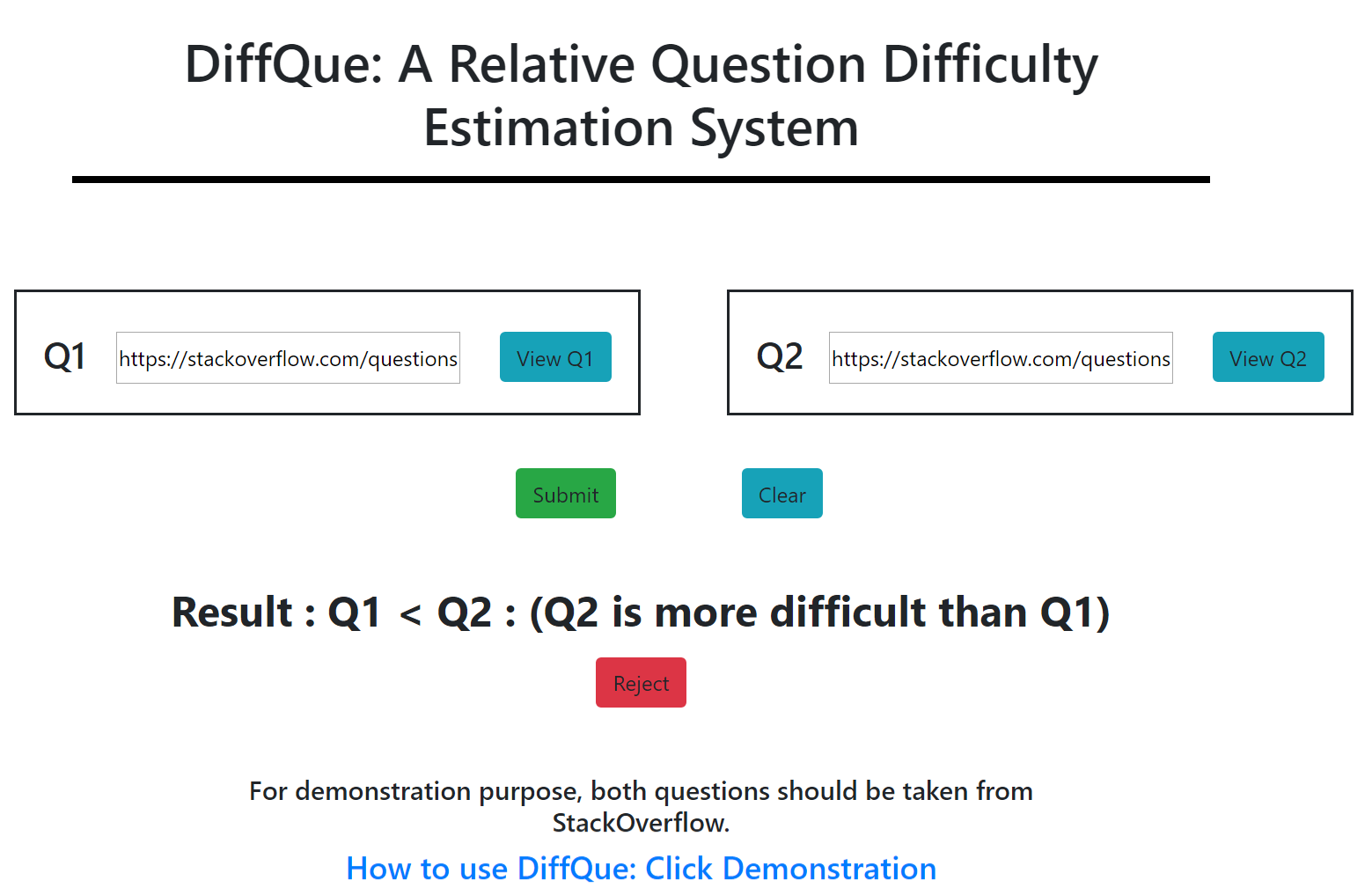}}
\caption{User interface of \system.}\label{fig:system}
\end{figure}

We have designed an experimental version of \system. Figure \ref{fig:system} shows the user interface of \system. Upon entering into the site, users need to put the link of two questions and press the `Submit' button. The system will then show which question is more difficult. The current implementation only accepts questions from Stack Overflow. However, it will be further extended for other (and across) CQA services.

\system~also supports {\bf online learning}. There is a `Reject' button associated with answer \system~produces, upon clicking of which the feedback will be forwarded to the back-end server. \system\ has the capability to be trained incrementally with the feedback provided by the users \cite{shilton2005incremental}. However, special care has been taken to make the learning process robust by ignoring spurious feedback. An attacker may want to pollute \system~ by injecting wrong feedback. \system~handles these spurious feedback by first checking the confidence of the current model on labels associated with the feedback, and ignoring them if the confidence is more than a pre-selected threshold (currently set as 0.75). This makes \system~more robust under adversarial attacks.

\section{Estimating Overall difficulty}
One may further be interested to assign a global difficulty score for each question, instead of predicting which one is more difficult between two questions. In this case,  questions could be labeled as several difficulty levels (e.g., easy, medium, hard). 
We argue  that solving this problem is computationally challenging as it may lack sufficient training samples, and it also needs an understanding of the difficulty levels of every pair of questions. Even if we design such system, its evaluation  would be extremely challenging -- it is relatively easy for a human annotator to find the difficult question from a pair compared to judging the overall difficulty score of each question.  

However, here we attempt to design such system which can label each question as `easy', `medium' or `hard'. Our hypothetical system would construct the complete network (ignoring the directionality) such that every node pair is connected by an undirected edge, and the directionality of every edge is predicted by \system. However, for a complete network with $n$ nodes, there are $\Comb{n}{2}$ edges which, for a large value of $n$, would require huge memory to store.  We therefore approximate this procedure by picking a random pair of nodes at a time and predicting the directionality of the edge connecting them using \system. We repeat this process $9,100,000$ times to construct a semi-complete network as permitted by the memory limitation of subsequent computations. We then compute \textbf{PageRank} of nodes on the constructed network.

Estimation of difficulty level based on PageRank value requires suitable thresholding. For this, we chose $400$ questions randomly from SO3 dataset and asked human annotators to annotate their difficulty level. The inter-annotator agreement using Fleiss' kappa is $0.56$; such a low value once again implies that even human annotators showed strong disagreement in labeling global difficulty level of questions. However, for further processing, we took $300$ such questions for which at least two annotators agreed on the difficulty level. Experiments are conducted using 10-fold cross validation where the training set is used for determining the thresholds. Table \ref{tab:overall} shows that \system~outperforms other baselines for all difficulty levels.

\begin{table}[!h]
\centering
\begin{tabular}{|l|c|c|c|c|c|c|}
\hline
\multirow{2}{*}{{\bf Model}} & \multicolumn{2}{c|}{{\bf Easy}} & \multicolumn{2}{c|}{{\bf Medium}} & \multicolumn{2}{c|}{{\bf Hard}} \\\cline{2-7}
      & {\bf P} & {\bf R} & {\bf P} & {\bf R} & {\bf P} & {\bf R}\\\hline
{\bf RCM} & 0.33 & 0.27 & 0.64 & 0.65& 0.50 & 0.23 \\
{\bf Trueskill} & 0.41 & 0.24 & 0.61 & 0.74& 0.32 & 0.28\\
{\bf \system} & 0.52 & 0.41 & 0.70 & 0.92 & 0.73 & 0.33 \\\hline
\end{tabular}
\caption{Accuracy (P: Precision, R: Recall) of completing models for estimating overall difficulty of questions on the SO3 dataset. Accuracy is reported for three levels of difficulty.}\label{tab:overall}
\vspace{-3mm}
\end{table}

\section{Conclusion}


In this paper, we proposed \system~to address the problem of estimating relative difficulty of a pair of questions in CQA services. \system~leverages a novel network structure and  estimates the relative difficulty of questions by running a supervised edge directionality prediction model. \system~turned out to be highly efficient than four state-of-the-art baselines w.r.t. the accuracy, robustness and capability of handling cold state problem. \system\ can further be used to obtain an overall ranking of all questions based on the difficulty level. We have made the code and dataset available for reproducibility. 

\section*{Acknowledgement}
The work was partially funded by DST, India (ECR/2017/00l691) and Ramanujan Fellowship. The authors also acknowledge the support of the Infosys Centre of AI, IIIT-Delhi, India.

\bibliographystyle{ACM-Reference-Format}
\bibliography{sample-bibliography}

\if{0}

\section{Supplementary Materials}

\section{Introduction}

Programmers these days often rely on various community-powered platforms such as Stack Overflow, MathWorks, Stack Exchange etc. -- also known as Community Question Answering (CQA) services to resolve their queries. A user posts a question/query which later receives multiple responses. The user can then choose the best answer (and mark it as `accepted answer') out of all the responses. Such platforms have recently gained huge attention due to various features such as quick response from the contributers, quick access to the experts of different topics, succinct explanation etc.  For instance, In August 2010, Stack Overflow accommodated $300k$ users and $833k$ questions; these numbers currently jumped to $8.3m$ users and $15m$ questions posted\footnote{\url{https://stackexchange.com/sites/}}. This in turn provides tremendous opportunity to the researchers to consider such CQA services as large knowledge bases to solve various interesting problems  \cite{cscm4,msr-paper,chinese-paper}.  



Plenty of research has been conducted on CQA services, including  question search \cite{sigir1}, code search \cite{anne1}, software development assistance \cite{rankingcrowdknowledge}, question recommendation \cite{recsys4} etc. A significant amount of study is done in estimating user expertise \cite{sigir2,www1}, recommending tags \cite{icsme1}, developing automated techniques to assist editing a post \cite{cscw1} etc. Studies are also conducted to help developers mapping their queries to the required code snippets \cite{nlp2code}.


{\bf Problem Definition and Motivation:} In this paper, {\em we attempt to automatically estimate the relative difficulty of a question among a given pair of questions posted on CQA services}. Such a system would help expert users to retrieve questions with desired difficulty, hence making best use of time and knowledge. 
It can also assist to prepare a knowledge base of questions with varying level of difficulty. Academicians can use this system to set up their question papers on a particular topic.
%
However, computing difficulty of a question is a challenging task as we cannot rely only on textual content or reputation of users. High reputation of a user does not always imply that his/her posted questions will be difficult.
Similarly, a question with embedded code (maybe with some obfuscation) does not always indicate the difficulty level. Therefore, we need to find a novel solution which uses different features of a question and interaction among various users to learn characteristics of the question and estimates the difficulty level. 

\begin{figure}[!t]
\centering
\includegraphics[width=\columnwidth]{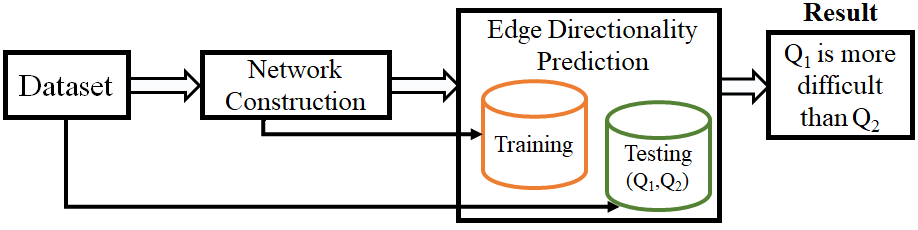}
\vspace{-5mm}
\caption{\system~framework.}\label{fig:framework}
\vspace{-5mm}
\end{figure}

\begin{table}[!t]
\centering
\begin{tabular}{|c|cccc|}
\hline
\multirow{2}{*}{{\bf Method}} & \multicolumn{4}{|c|}{{\bf Dataset}}\\\cline{2-5}
 & {\bf SO1} & {\bf SO2} & {\bf SO3} & {\bf MSE}\\\hline
 RCM &55.60 & 57.64& 50.72& 55.34\\
Tureskill & 55.98  & 52.8& 52.67 & 57.62 \\
PageRank & 47.26 & 49.36 & 50.09 & 50.79  \\
HITS & 45.95&54.6 & 49.15& 54.91 \\\hline
\system & {\bf 72.24} & {\bf 70.56} & {\bf 76.39} & {\bf 70.23} \\\hline
\end{tabular}
\caption{F\textsubscript{1} score of the competing methods on four different datasets -- SO1, SO2, SO3, MSE (Baseline: RCM, Trueskill, PageRank and HITS) (Section \ref{sec:result} for more details). \system~outperforms other baselines across all the datasets.}\label{tab:summary}
\vspace{-3mm}
\end{table}

{\bf Proposed Framework:} In this paper, we propose \system, a relative difficulty estimation system for a pair of questions that leverages a novel (directed and temporal) network structure generated from the user interactions through answering the posted questions on CQA services. \system~follows a two-stage operation (Figure \ref{fig:framework}). In the first stage, it constructs a network whose nodes correspond to the questions posted, and edges are formed  based on a set of novel hypotheses (which are statistically validated) derived from the temporal information and user interaction available in CQA services.  In the second stage, \system~ maps the `relative question  difficulty estimation' problem to an `edge directionality prediction' problem. It extracts features from the user metadata, network structure and textual content and runs a supervised classifier to estimate relatively difficult question among a question pair.  In short, \system~  systematically captures three fundamental dimensions of a CQA service -- user metadata, textual content and temporal information.

{\bf Summary of the Results:}
We evaluate the performance of \system~on two CQA platforms -- Stack Overflow (which is further divided into three parts based on the time) and Mathematics Stack Exchange. We compare \system~with two other state-of-the-art techniques -- Trueskill \cite{msr-paper} and PageRank \cite{chinese-paper} along with another baseline (HITS) we propose. All these baselines leverage some kind of network structure. Experimental results (Table \ref{tab:summary}) on human annotated data show that \system~outperforms all the baselines by a significant margin -- \system~achieves  $72.89\%$ F\textsubscript{1} score ({\em resp.} ?? AUC) on average across all the datasets, which is ??\% ({\em resp.} ??\%) higher than the best baseline. We statistically validate our network construction model and show that if other baselines leverage our network structure instead of their own, they could improve the accuracy significantly (on average ??\%, ??\% and ??\% improvement for Trueskill, PageRank and HITS respectively) compared to their proposed  system configuration. Further analysis on the performance of \system~reveals that \system~is -- (i) robust to noise, (ii) flexible to domain adaptation, and (iii) resilient to cold start problem.

{\bf Contribution of the Paper:} In short, the major contributions of the paper are four-fold:
\begin{itemize}
\item We propose a novel network construction technique by leveraging the user interactions and temporal information available in CQA services. The network provides a relative ordering of pairs of questions based on the level of difficulty. We also show that the baselines could improve their performance if they  use our network instead of theirs. 

\item We map the problem of `relative difficulty estimation of questions' to an `edge directionality prediction' problem, which, to our knowledge, is the first attempt of this kind to solve this problem. Our proposed method utilizes three fundamental dimensions of a CQA service -- user information, textual content and time information. 

\item \system~truns out to be superior to the state-of-the-art methods -- It not only beats the other baselines significantly, but also is appropriately robust to noise and cold start problem.  

\item As a by-product of the study, we generated huge CQA datasets and manually annotated a set of question pairs based on the difficult level. This may become a valuable resource for the research community.

\end{itemize}
{\bf For the sake of reproducible research, we have made the code publicly available at \url{????}}. \footnote{The datasets will be released upon acceptance of the paper.}

\section{Related Work}
Recently, we have seen expansion of various CQA services, including Stack Overflow, Quora, Reddit etc. There are number of studies that involve these CQA services. Here we organize the literature review in two parts: general Study involving CQA and studies on difficulty ranking of questions.	 

\subsection{General Study involving CQA}
CQA has grown tremendously in size, thus providing huge databases to the research community.  \cite{codeexample, rwc2} mined the Stack Overflow data and provided ways to identify helpful answers. \cite{recomm} recommended questions available in CQA to experts based on their domain expertise. \cite{kdd1} observed user's lifetime in CQA services and found that it does not follow  exponential distributions. There are many studies that analyzed temporal effect in CQA sites \cite{ieee1,wsdm1,icwsm2}. \cite{icwsm2} showed that there are two strategies to develop online authority: progressively developing reputation or by using prior acquired fame. \cite{ieee1} found online user activities time-dependent, and proposed question routing mechanism for timely answer. 
\cite{temporaljournal} found three parameters to capture temporal significance and to personalize PageRank: frequency factor (trustworthy users are often active), built-up time length factor (the longer the time between registration of user and link creation, the more trustworthy the user) and similarity factor (the pattern in which two users add links).

\subsection{Difficulty Ranking of Questions}\label{sec:rw:qd}
\cite{chinese-paper, msr-paper} attempted to solve the problem of finding the difficulty of tasks taking the underlying network structure into account. \cite{chinese-paper} constructed a network in such a way that if user $A$ answers questions $Q_1$ and $Q_2$, among which only the answer of $Q_1$ is accepted, then there will be a  directed edge from $Q_2$ to $Q_1$. Following this, they used {\bf PageRank} as a proxy of the difficulty of a question.
\cite{msr-paper} created a user network where each question $Q$ is treated as a pseudo user U$_{Q}$. It considers four competitions: one between pseudo user U$_{Q}$ and asker U$_{a}$, one between pseudo user U$_{Q}$ and the best answerer U$_{b}$, one between the best answerer U$_{b}$ and asker U$_{a}$ and one between the best answerer U$_{b}$ and each of the non-best answers of the question $Q$. It then uses {\bf Trueskill}, a Bayesian skill rating model. Both of \cite{chinese-paper} and \cite{msr-paper} neither used textual content nor temporal effect which play a crucial role in determining the importance of questions over time. \cite{regularised} used the same graph structure as \cite{msr-paper} and proposed Regularized Competition Model ({\bf RCM}) to capture the significance of difficulty. It forms $\theta \in \mathcal{R}^{M+N}$, denoting the `expert score' of pseudo users -- initial $M$ entries are expertise of users while further $N$ are difficulty of questions. For each of the competitions, $x_k$ vector is formed where ${x_i}\textsuperscript{k}$ = 1, ${x_j}\textsuperscript{k}$ = -1 and $y_k$ = 1 if $i$ wins over $j$, else $y_k$ = -1. The algorithm starts at initial $\theta$ and proceeds towards negative subgradient, $\theta_{t+1}$ = $\theta_t$ - $\gamma_t$*$\triangledown$$\mathcal{L}$($\theta_t$), where $\triangledown$$\mathcal{L}$($\theta_t$) is the subgradient and $\gamma_t$ is set as 0.001. Further, classification of examination questions according to Bloom's Taxonomy \cite{rccst1} tries to form an intuition of difficulty of questions but it is not very helpful in case of CQA services, where there are many users, and each of them has their ways of expressing the questions, or sometimes there is code attached with question where normal string matching can trivialize the problem. We consider all these methods as baselines for \system.


\section{Datasets} \label{sec:dataset}

We collected questions and answers from two different CQA services -- (i) Stack Overflow\footnote{\url{https://stackoverflow.com/}} (SO) and Mathematics Stack  Exchange\footnote{\url{https://math.stackexchange.com/}} (MSE), both of which are extensively used  by developers or mathematicians to get their query resolved. The former dataset was further divided into three parts -- SO1, SO2 and SO3 based on time of posting the questions. A brief description of the datasets are presented below (see Table \ref{tab:dataset}).

\subsection{Stack Overflow (SO) Dataset}
This dataset contains all the questions and the answers of Stack Overflow till August 27, 2017, accommodating more than $10m$ questions (70\% of which are answered) and more than $20m$ answers in total. In this work, We filtered only ``Java-related questions''. Further we divided the selected questions and answers into three parts based on the time of posting the questions -- questions posted in first 2 years, middle 2 years and last 2 years form the {\bf SO1}, {\bf SO2} and {\bf SO3} datasets respectively. User metadata was also available along with the questions and answers. The intuition behind dividing the entire data into three parts is to capture the dynamics of user interaction in different time points, which may change over time. 

\subsection{Mathematics Stack exchange (MSE) Dataset}
This dataset contains all the questions posted on MSE till August 27, 2017 and their answers. There are about $800K$ questions and about a million answers. About 75\% questions are answered. Here we only extracted questions related to the following topics: Probability, Permutation, Inclusion-Exclusion and Combination\footnote{The topics were chosen based on the expertise of the experts who further annotated the datasets.}. The questions and answers belonging to these topics are filtered from the entire database to prepare our dataset. The number of users in the chosen sample is $47,470$. Unlike SO, we did not divide MSE into different parts because total number of questions present in MSE is very  less compared to SO (less than even a single part of SO). Further division have reduced the size of the dataset significantly. 

Both these datasets are available as dump in xml format. Each post (question/answer) comes with other metadata: upvotes/downvotes, posting time, answer accepted or not, tag(s) specifying the topics of the question etc. Similarly, the metadata of users include UserID, registration time and reputation. Unlike other baselines \cite{msr-paper,chinese-paper}, we also utilize these attributes in \system.



\begin{table}[t!]
  \centering
    \resizebox{\columnwidth}{!}{%
    \begin{tabular}{c|c|c|c|c}
     \hline
 {\bf Dataset} & {\bf \# questions} & {\bf \# answers} & {\bf \# users} & {\bf Time period} \\
 \hline
 SO1 & 100,000 & 289,702 & 60,443 & Aug'08 --  Dec'10 \\ 

 SO2 &  342,450  & 603,402 &  179,827 & Jan'12 -- Dec'13 \\
 SO3 & 440,464 & 535,416 & 274,421 & Aug'15 -- Aug'17 \\\hline

 MSE & 92,686 & 119,754 & 47,470 & July'10 -- Aug'17  \\ 
 \hline
  \end{tabular}
  }
     \newline
      \caption{Statistics of the datasets.}
      \label{tab:dataset}
      \end{table}

\section{\system: Our Proposed Framework}
\system~first maps a given CQA data to a directed and longitudinal network where each node corresponds to a question and an edge pointing from one question to another question indicates that the latter question is harder than the former one. Once the network is constructed, it trains a edge directionality prediction model on the given network and predicts the directionality of an edge connecting two given questions under inspection. The framework of \system~is shown in Figure \ref{fig:framework}. Rest of the section elaborates each component of \system.

\subsection{Network Construction}\label{sec:networkconst}
\system~models the entire dataset as a longitudinal\footnote{Nodes are arranged based on their creation time.} and directed network $G=(V,E)$, where $V$ indicates a set of vertices and each vertex corresponds to a question; $E$ is a set of edges. Each edge can be of one of the following three types mentioned below.

\vspace{1mm}
\noindent\fbox{%
    \parbox{\columnwidth}{%
{\bf Nomenclature:} Throughout the paper, we will assume that {\bf Bob} has correctly answered  {\bf Robin}'s question, and therefore Bob has more expertise than Robin.}}\vspace{2mm}

\noindent {\bf \underline{Edge Type 1:}} An expert on a certain topic does not post trivial questions on CQA sites. Moreover, s/he answers those questions which s/he has expertise on. We capture these two notions in Hypothesis \ref{hyp1}. 

\begin{hyp}\label{hyp1}
If Bob correctly answers question $Q$ asked by Robin, then the questions asked by Bob later will be considered more difficult than $Q$.
\end{hyp}
The `correctness' of an answer is determined by the `acceptance' status of the answer or positive number of upvotes (provided by CQA services).

Let $Q_R$ be the question Robin posted at time $T_R$ and Bob answers $Q_R$. Bob later asks $n$ questions, namely $Q_{B_1}, Q_{B_2}, \cdots, Q_{B_n}$ at $T_{B_1}, T_{B_2}, \cdots, T_{B_n}$ respectively.  Then the difficulty level of each $Q_{B_i}$, denoted by $diff(Q_{B_i})$ will be more than that of  $Q_{R}$, i.e., $diff(Q_{B_i})\geq diff(Q_{R}), \forall i$. The intuition behind this hypothesis is as follows -- since Bob correctly answered $Q_R$, Bob is assumed to have more expertise than the expertise required to answer $Q_R$. Therefore, the questions that Bob will ask later may need more expertise than that of  $Q_R$. 

We use this hypothesis to draw edges from an easy question to a difficult question as follows: an edge $e=\langle x,y \rangle \in E$ of type 1 indicates that $y$ is more difficult than $x$ according to Hypothesis \ref{hyp1}. Moreover, each such edge will always be a forward edge, i.e., a question asked earlier  will point to the question asked later. Edge $\langle Q_{R_2}, Q_{B_3} \rangle$ in Figure \ref{fig:example} is of type 1.

One may argue that the {\em answering time} is also important in determining type 1 edges -- if at $T_1$ Bob answers $Q_R$ which has been asked by Robin at $T_{Q_R}$ (where $T_1>T_{Q_R}$), we should consider all Bob's questions posted after $T_{1}$ (rather than $T_{Q_R}$) to be difficult than $Q_R$. However, in this paper we do not consider  answering time separately and assume question and answering times to be the same since the time difference between posting the question and answering the  question, i.e.,  $T_1-T_{Q_R}$ seems to be negligible across the datasets (on average 17.21, 11.33, 7.12, 11.34 days for SO1, SO2, SO3 and MSE respectively). \\

\noindent \textbf{\underline{Edge Type 2:}} It is worth noting that an edge of type 1 only assumes Bob's questions to be difficult which {\em will be} be posted later. It does not take into account the fact that all Bob's contemporary questions (posted very recently in the past) may be difficult than Robin's current question; even if the former questions may be posted slightly before the latter question. We capture this notion in type 2 edges using Hypothesis \ref{hyp2}.

\begin{hyp}\label{hyp2}
If Bob correctly answers Robin's question $Q$, then Bob's very recent posted questions will be more difficult than $Q$.
\end{hyp}

Let $Q_{B_1}, Q_{B_2}, \cdots$ be the questions posted by Bob at $T_{B_1}, T_{B_2}, \cdots$ respectively. Bob has answered Robin's question $Q_R$ posted at $T_{R}$ and $T_{R}\geq T_{B_i}, \forall i$. However the difference between $T_{R}$ and $T_{B_i}$ is significantly less, i.e., $T_R-T_{B_i}\leq \delta_t, \forall i$ which indicates that all these questions are contemporary. According to Hypothesis \ref{hyp2}, $diff(Q_{B_i})\geq diff(Q_R)$.

We use this hypothesis to draw edges from easy question to difficult question as follows: an edge $e=\langle x,y \rangle \in E$  of type 2 indicates that $y$ is more difficult than $x$ and $y$ was posted within $\delta_t$ times before the posting of $x$. Note that each such edge is backward edge, i.e., a question asked later may point to the question asked earlier. Edge $\langle Q_{R_2}, Q_{B_2} \rangle$ in Figure \ref{fig:example} is of type 2. \\

\noindent{\bf \underline{Edge Type 3:}} We further consider questions those are posted by a single user over a time in Hypothesis \ref{hyp3}. 

\begin{hyp}\label{hyp3}
Over the time, a user's expertise will increase, and thus the questions that s/he will ask will keep becoming difficult. 
\end{hyp}

Let $Q_{B_1}, Q_{B_2}, \cdots$ be the questions posted by Bob\footnote{Same hypothesis can be applied to any user (Bob/Robin).} at $T_{B_1}, T_{B_2}, \cdots$ , where $T_{B_{i+1}}> T_{B_i}, \forall i$. Then $diff(Q_{B_{i+1}})>diff(Q_{B_i}) \forall i$. The underlying idea is that as the time progresses, a user gradually becomes more efficient and acquires more expertise on a particular topic. Therefore, it is more likely that s/he will post questions which will be more difficult than his/her previous questions. 

We use this hypothesis to draw an edge from an easy question to a difficult question as follows: an edge $e=\langle x,y \rangle \in E$ of type 3 indicates that: (i) both $x$ and $y$ were posted by the same user, (ii) $y$ was posted after $x$ and there was no question posted by the user in between the posting of $x$ and $y$ (i.e., $T_y>T_x \& \nexists z: T_y>T_z>T_x$) , (iii) $y$ is more difficult than $x$. Note that each such edge will also be a forward edge. Edge $\langle Q_{B_2}, Q_{B_3} \rangle$ in Figure \ref{fig:example} is of type 3, but $Q_{B_2}$ and $Q_{B_4}$ should not be connected according to this hypothesis.

Note that an edge can be formed by more than one of these hypotheses. However, we only keep one instance of such edge in the final network. All these hypotheses are statistically significantly (see Section \ref{sec:hypothesis}). The number of edges of each type in the datasets is shown in Table \ref{edges}. \\

\noindent {\bf \underline{Parameters of the network model:}} There are two parameters to construct the network:
\begin{itemize}
\item {\bf Time of posting an answer:} Instead of considering the absolute time as the posting time of a question, we divide the entire timespan present in a dataset into 2-week buckets, and assume that if two questions are posted within a bucket, their posting time is same. However, we further vary bucket size from 1 week to 6 weeks and observe that 2-week bucket size produces the best results (see Table \ref{tab:parameter}).\todo{The context of this discussion is not clear: DONE. Need to tell how actually it effects?}

\item {\bf Recency of questions for type 2 edges:} In edge type 2, we have introduced $\delta_t$ quantifying the recency of Bob's question $Q_{B_i}$. $Q_{B_i}$ has been asked `very recently' before Robin's question $Q_{R}$ which Bob has correctly answered.
We vary $\delta_t$ from $1$ to $5$ \todo{report the result} and observe that $\delta_t=1$ produces the best result.
\end{itemize}

\begin{table}
  \centering
    \begin{tabular}{c|c|c|c}
     \hline
          \multirow{2}{*}{{\bf Dataset}} & \multicolumn{3}{|c}{{\bf Edge count}} \\  \cline{2-4}
          & {\bf Edge Type 1} & {\bf Edge type 2} & {\bf Edge Type 3} \\ 
         \hline
         SO1 & 749,757 & 61,209 & 133,243 \\
         SO2 & 1,168,490 & 101,196  & 392,743 \\
         SO3 & 556,511 & 42,010  & 319,222  \\
         MSE & 224,058 & 10,124 & 89,996 \\
         \hline
  \end{tabular}
     \newline
      \caption{Number of edges of each type in different datasets.}
      \label{edges}
      \vspace{-5mm}
      \end{table}

\begin{example}
Let us consider that Robin has asked three questions, $Q_{R_1}$, $Q_{R_2}$ and $Q_{R_3}$ at time $0$, $2$ and $4$ respectively, and Bob has answered $Q_{R_2}$. Bob has also asked four questions $Q_{B_1}$, $Q_{B_2}$, $Q_{B_3}$ and $Q_{B_4}$ at time $0$, $1$, $3$ and $4$ respectively. Figure \ref{fig:example} shows the corresponding network for this example.

\begin{figure}[t!]
    \centering
           \includegraphics[width=\columnwidth]{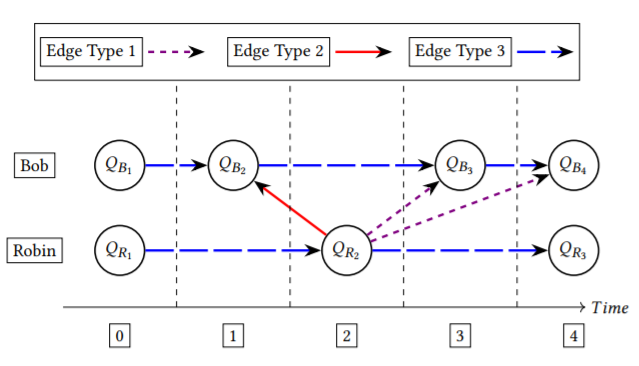}     
           \caption{A example depicting the network construction in \system. Here Bob has answered	 Robin's question $Q_{R_2}$.}
               \label{fig:example}
\end{figure}

\begin{table*}[!t]
\centering
\begin{tabular}{p{5.5cm}|p{5.5cm}|p{5.5cm}}
\hline
{\bf Hypothesis 1} & {\bf Hypothesis 2} & {\bf Hypothesis 3}\\\hline
\underline{Sample 1} & \underline{Sample 2} & \underline{Sample 3}\\
{\bf (Q1)} Write a program to generate all elements of power set? & {\bf (Q1)} Given an array of size n, give a deterministic algorithm (not quick sort) which uses O(1) space (not median of medians) and find the K\'th smallest item. & {\bf (Q1)} Given an array of integers (each <= $10^{6}$) what is the fastest way to find the sum of powers of prime factors of each integer?\\
{\bf (Q2)} How can one iterate over the elements of a stack starting from the top and going down without using any additional memory. The default iterator() goes from bottom to top. For Deque, there is a descendingIterator. Is there anything similar to this for a stack. If this is not possible which other Java data structures offer the functionality of a stack with the ability to iterate it backwards? 
& {\bf (Q2)} Transform a large file where each line is of the form: b d, where b and d are numbers. Change it from:
b -1 to b 1, where b should remain unchanged. There is a huge file and a way is required to achieve this using, say, sed or a similar tool?
& {\bf (Q2)} Given an array of N positive elements, one has to perform M operations on this array. In each operation a subarray (contiguous) of length W is to be selected and increased by 1. Each element of the array can be increased at most K times. One has to perform these operations such that the minimum element in the array is maximized. Only one scan is required.\\\hline
{\bf Response:} $diff$(Q2)>$diff$(Q1)? Yes: 75\%  & {\bf Response:} $diff$(Q2)>$diff$(Q1)? Yes: 75\% & {\bf Response:} $diff$(Q2)>$diff$(Q1)? Yes: 80\% \\\hline\hline
\underline{Random sample 1} & \underline{Random sample 2} & \underline{Random sample 3}\\ 
{\bf (Q1)} If an object implements the Map interface in Java and one wish to iterate over every pair contained within it, what is the most efficient way of going through the map? & {\bf (Q1)} Given a string s of length n, find the longest string t that occurs both forward and backward in s. e.g, s = yabcxqcbaz, then return t = abc or t = cba. Can this be done in linear time using suffix tree? & {\bf (Q1)} What happens at compile and runtime when concatenating an empty string in Java?
\\
{\bf (Q2)} How to format a number in Java? Should the number be formatted before rounding? 
& {\bf (Q2)} Given two sets A and B, what is the algorithm used to find their union, and what is it's running time?
& {\bf (Q2)} What is the simplest way to convert a Java string from all caps (words separated by underscores) to CamelCase (no word separators)?
\\\hline
{\bf Response:} $diff$(Q2)>$diff$(Q1)? Yes: 20\%  & {\bf Response:} $diff$(Q2)>$diff$(Q1)? Yes: 35\% & {\bf Response:} $diff$(Q2)>$diff$(Q1)? Yes: 30\% \\\hline\hline

\end{tabular}
\caption{One example question pair for each sample and its corresponding human response.} 
\label{surveyy}
\end{table*}

\end{example}

\subsection{Hypothesis Testing}\label{sec:hypothesis}
In this section, we present a thorough survey and show that three hypotheses behind our edge construction framework are statistically significant.
For this, we prepared $6$ sets of edge samples, each two generated for each hypothesis as follows:
\begin{enumerate*}
    \item Sample 1: Choose $20$ edges of type 1 randomly from the network. 
    \item Random Sample 1: Randomly select $20$ pairs of questions which obey the time constraint mentioned in Hypothesis \ref{hyp1}.
    \item Sample 2: Choose $20$ edges of type 2 randomly from the network.
    \item Random Sample 2: Randomly select $20$ pairs of questions such that they follow the recency mentioned in Hypothesis \ref{hyp2}.
    \item Sample 3: Choose $20$ edges of type 3 randomly from the network.
    \item Random Sample 3: Randomly select $20$ pairs of questions such that they follow the time constraint mentioned in Hypothesis \ref{hyp3}.
\end{enumerate*}

The survey was conducted with $20$ human annotator of age between 25-35, who are experts on java and math-related domains. For a question pair ($Q_x,Q_y$) (i.e., an edge $\langle x,y \rangle$) taken from each sample, we hypothesize that $diff(Q_y)>diff(Q_x)$. Null hypothesis rejects our hypothesis. Given a question pair, we asked each annotator to mark $1$ if our hypothesis holds; otherwise mark $0$. One example question pair for each sample and the corresponding human response statistics (percentage of annotators accepted our hypothesis) are shown in Table \ref{surveyy}. Table \ref{pvalue} shows that for each hypothesis the average number of annotators who accepted the hypothesis is higher for the sample taken from our network as compared to that for its corresponding random sample (the difference is also statistically significant as $p<0.01$, Kolmogorov-Smirnov test). This result therefore justifies our edge construction mechanism and indicates that all our hypotheses are realistic.

\begin{table}[!h]
  \centering
    \begin{tabular}{c|c|c|c}
     \hline
 {\bf Hypothesis} & {\bf Original sample} & {\bf Random sample} & {\bf p-value}  \\\hline
 H1 &  13.25 & 10.44 & p \textless 0.01 \\

 H2 & 12.12 & 10.11 & p \textless 0.01 \\
 
 H3  & 15.4  & 10.66 & p \textless 0.01 \\  \hline
  \end{tabular}
     \newline
      \caption{Average number of annotators who accepted the hypotheses, and the p-value indicating the significance of our hypotheses w.r.t the null hypothesis.}
      \label{pvalue}
    \vspace{-4mm}
      \end{table}

\subsection{Edge Directionality Prediction Problem}
Once the network is constructed, \system~considers the `relative question difficulty estimation' problem as an `edge directionality prediction' problem. Since an edge connecting two questions in a network points to the difficult question from the easy question, given a pair of questions with unknown difficulty level, the task boils down to the prediction of an directed edge which (virtually) connects these two questions.

Although research on like prediction is vast in network science domain \cite{LU20111150}, the problem of edge directionality prediction is relatively less studied. Guo et al.  \cite{Guolink-prediction} proposed a ranking-based method that combines both local indicators and global hierarchical structures of networks for predicting the direction of edge.  Soundarajan and Hopcroft \cite{Soundarajan} treated this problem as a supervised learning problem. They calculated various features of each known links based on its position in the network, and used SVM to predict the unknown directions of edges. Wang et al. \cite{Wang_link_prediction} proposed local directed path that solves the information loss problem in sparse network and makes the method effective and robust.


We also consider `edge directionality prediction problem' as a supervised learning problem. The pairs of questions, each of which is directly connected via edges in our network, form the training set.  
If the pair of questions under inspection are already connected directly via an edge in our network, the problem will be immediately solved by looking at the directionality of the edge. In our supervised model, given a  pair of questions ($a,b$) (one entity in the population), we first connect them by an edge if they are not connected and then determine the directionality of the edge. Our supervised model uses the following features which are broadly divided into three categories: (i) {\em network topology based} (F1-F6), (ii) {\em metadata based} (F7-F10), and (iii) {\em content based} (F11-F12) 

\begin{itemize}
    \item {\bf [F1] Leader Follower Ranking for node $a$:}  Guo et al. \cite{Guolink-prediction} used a ranking based approach to predict links in a network. At each step of the algorithm, for each node, the difference between node's in-degree and out-degree, $\gamma$ is computed to separate leaders (high-ranked nodes) from followers (low-ranked nodes). The algorithm uses $\alpha$, 0.65 in our case, specifying the proportion of leaders at each step. The whole network is partitioned into leaders and followers, and this process continues recursively on the further groups obtained. If at any time, number of nodes in any of the group, $\|V\| \leq  1/\alpha$, then ranking is done according to $\gamma$. During the merging of leaders and followers, followers are placed after leaders. Finally, any node ranked lower (less important) is predicted to have an edge to higher ranked nodes (more important). The ranking is normalized by the number of questions and then used as a feature. 
    \item {\bf [F2] Leader Follower Ranking for node $b$:} We use the similar strategy to measure the above rank for node $b$. 
    \item {\bf [F3] PageRank of nodes $a$:} It emphasizes the importance of a node, i.e. a probability distribution signifying if a random walker will arrive to that node.
We use PageRank to compute score of each node. However, we modify the initialization of PageRank by incorporating the weight of nodes -- let $A$ be the user asking the question $Q_A$ and the reputation (normalized to the maximum reputation of a user) of $A$ be $r_{A}$. Then the PageRank score of $Q_A$ is calculated as:
\begin{equation}
     PR(Q_A) ={(1-d)} {r_{A}} + d \sum_{Q_j \in N(Q_A)} \frac{PR (Q_j)}{Outdegree(Q_j)}
\end{equation}
Here, $N(Q_A)$ is the neighbors of $Q_A$ pointing to $A_A$; the damping factor $d$ is set to $0.85$.

\item {\bf [F4] PageRank of nodes $b$:} We similarly calculate the PageRank of node $b$ as mentioned above. 

    \item {\bf [F5] Degree of node $a$:} It is computed after considering the undirected version of network, i.e., number of nodes adjacent to a node.
    \item {\bf [F6] Degree of node $b$:} We calculate the degree of node $b$ similarly as mentioned above.    
 
    \item {\bf [F7] Posting time difference between $a$ and its accepted answer:} It signifies the difference between the posting time of $a$ ($T_a$) and its accepted answer $a'$ ($T_{a'}$), if any. If none of the answers is accepted, it is set to 1. However, instead of taking the direct time difference, we employ an exponential decay as:
    $1 - e^{-(T_{a'}-T_a)}$.   
The higher the difference between the posting time (implying that users have taken more times to answer the question), the  more the difficulty level of the question. 

    \item {\bf [F8] Posting time difference between $b$ and its accepted answer:} Similar score is computed for $b$ as mentioned above.

    \item {\bf [F9] Accepted answers of users who posted $a$ till $T_a$:} The more the number of accepted answers of the user asking question $a$ till $a$'s posting time $T_a$, the more the user is assumed to be an expert and the higher the difficulty level of $a$. We normalize this score by the maximum value among the users.
    \item {\bf [F10] Accepted answers of users who posted $b$ till $T_b$:} Similar score is measured for question $b$.

    \item {\bf [F11] Textual feature of $a$:} This factor takes into account the text present in $a$. The idea is that if a question is easy, its corresponding answer should be present in a particular passage of a text book. Therefore, for $a$ we first consider its accepted answer and check if the unigrams present in that answer are also available in different books.  For Java-related questions presents in SO1, SO2 and SO3, we consider the following books: (i) Core Java Volume I by Cay S. Horstmann \cite{Gvero:2013}, (ii) Core Java Volume II by Cay S. Horstmann \cite{jbook2},(iii) Java: The Complete Reference by Herbert Schildt \cite{jbook3}, (iv) OOP - Learn Object Oriented Thinking and Programming by Rudolf Pecinovsky \cite{jbook4}, and (v)  Object Oriented Programming using Java by Simon Kendal \cite{jbook5}. For Math-related questions present in MSE, we consider the following books: (i) Advanced Engineering Mathematics by Erwin Kreyszig \cite{mbook1}, (ii) Introduction to Probability and Statistics for Engineers and Scientists by Sheldon M. Ross \cite{mbook2}, (iii) Discrete Mathematics and Its Applications by Kenneth Rosen \cite{mbook3}, (iv) Higher Engineering mathematics by B.S. Grewal \cite{mbook4}, and (v) Advanced Engineering Mathematics  by K. A. Stroud \cite{mbook5}. For each type of questions (Java/Math), we merge its relevant books and create a single document. After several pre-processing (tokenization, stemming etc.) we divide the document into different paragraphs, each of which forms a passage. Then we measure TF-IDF based similarity measure between each passage and the accepted answer. Finally, we take the maximum similarity among all the passages. The intuition is that if an answer is straightforward, most of its tokens should be concentrated into one passage and the TF-IDF score would be higher as compared to the case where answer tokens are dispersed into multiple passages. If there is no accepted answer for a question, we consider this feature as $0$.  
    
\item {\bf [F12] Textual feature of $b$:}
We apply the same technique mentioned above and measure the textual feature of $b$.

\end{itemize}

In our supervised model, for each directed edge ($a,b$) ($b$ is more difficult than $a$), we consider ($a,b$) as an entity in the positive class (class 1) and ($b,a$) as an entity in the negative class (class 2). Therefore, in the training set the size of class 1 and class 2 will be same and equal to the number of directed edges in the overall network. We use different type of classifiers, namely SVM, Decision Tree, Naive Bayes, K Nearest neighbors (MLP) and  Multilayer Perceptron (MLP); among them SVM turns out to be the best model (Table \ref{tab:parameter}(c)).


\section{Experimental Results}\label{sec:result}
This section presents the performance of the competing methods. It starts by briefly describing the human annotation for test set generation, followed by the baseline methods. It then thoroughly elaborates the parameter selection process for \system, comparative evaluation of the competing methods and other additional evaluations to check the robustness against noise and domain switching.  All computations were carried out on single Ubuntu machine with $32$ GB RAM and $2.7$GHz CPU with $12$ cores.

\subsection{Test Set Generation}
For each of the datasets mentioned in Section \ref{sec:dataset}, we prepared a test set for evaluating the competing models. The test set was generated as follows:
\begin{itemize}
\item We randomly selected many pairs of questions from each dataset.
\item Three experts\footnote{All of them also helped us validating our hypotheses  as discussed in Section \ref{sec:hypothesis}.} independently annotated each pair and marked their difficulty relationship. 
\item Finally, we selected $300$ such pairs of questions from each dataset such that all three annotators agreed upon their annotations. 
\end{itemize}

The inter-annotator agreement using Cohen's kappa is $0.84$. We further use total $1,200$ such annotations for comparative analysis. 


\subsection{Baseline Models and Evaluation Measures}
We consider three baseline methods described below; first two are existing methods and the last one is designed by us:\\
$\bullet$ {\bf RCM:} The Regularized Competition Model proposed by Wang et al. \cite{regularised} (Section \ref{sec:rw:qd}).\\
$\bullet$ {\bf Trueskill:} The approach proposed by Liu et al. \cite{msr-paper} (Section \ref{sec:rw:qd}).\\
$\bullet$ {\bf PageRank:} The approach proposed by Yang et al. \cite{chinese-paper} (Section \ref{sec:rw:qd}).\\
$\bullet$ {\bf HITS:} We further propose a new baseline as follows -- we run HITS algorithm \cite{hits} on our network. We computed authority scores for every question and rank the questions globally based on their authoritativeness. We used authority of a question instead of hub score to evaluate the  node (question) in determining the difficulty rather than the links. Now given a pair of questions, we mark the one as more difficult whose authoritative score is higher. 

We measure the accuracy of the methods based on F\textsubscript{1} score (in \%) and Area under the ROC curve (AUC).

\subsection{Parameter Selection for \system}
There are three important parameters of \system: (i) bucket size for determining question posting time, (ii) recency of questions for type 2 edges ($\delta_t$) and (iii) classifier for link prediction. Table \ref{tab:parameter} shows the  F\textsubscript{1} score of \system~by varying the parameter values on SO3 dataset\footnote{The pattern was same for the other datasets and therefore not reported here.}. We vary the bucket size from $1$ to $6$ and observe that $2$-week bucket interval produces the best result. $\delta_t$ is varied from $1$ to $6$ and $\delta_t=1$ gives the highest accuracy. We use five classifiers: SVM, Decision Tree (DT), Naive Bayes (NB), K Nearest Neighbors (KNN) and Multilayer Perceptron (MLP) with suitable hyper-parameter optimization; among them SVM (with RBF kernel) produces the best accuracy.  Therefore, unless otherwise stated, {\bf we use the following parameters as default for \system: bucket size=2, $\delta_t=1$ and SVM classifier}.

\begin{table}
\centering
\begin{tabular}{c|c|c|c|c|c}
\multicolumn{6}{c}{{\bf (a) Bucket size}}\\\hline
1 & 2 & 3 & 4 & 5 & 6\\\hline
70.1 & {\bf 76.39} & 71.1 & 68.3 & 70.1 & 60.4 \\\hline
\end{tabular}

\begin{tabular}{c|c|c|c|c|c}
\multicolumn{6}{c}{{\bf (b) $\delta_t$}}\\\hline
1 & 2 & 3 & 4 & 5 & 6\\\hline
{\bf 76.39} & 76.11 & 75.88 & 75.88 & 75.91 & 76.2  \\\hline
\end{tabular}

\begin{tabular}{c|c|c|c|c}
\multicolumn{5}{c}{{\bf (c) Classifier}}\\\hline
SVM & DT & NB & KNN & MLP\\\hline
{\bf 76.39}  & 63.9  & 75.2 & 69.1  & 75.4\\\hline
\end{tabular} 
\caption{F\textsubscript{1} score of \system~ with different parameter combination on SO3 dataset. For each parameter, we vary its value keeping the other parameters default.  }\label{tab:parameter}
\vspace{-5mm}
\end{table}

\begin{table*}[t!]
\centering
\begin{tabular}{|c|c|c|c|c|c|c||c|c|c|c|c|}
\hline
   \multirow{2}{*}{{\bf Dataset}} & {\bf Underlying} & \multicolumn{5}{|c||}{{\bf F\textsubscript{1} score (\%)}} &  \multicolumn{5}{|c|}{{\bf AUC}} \\ \cline{3-12}
          & {\bf Network} & {\bf RCM} & {\bf Trueskill} & {\bf PageRank} & {\bf HITS} & {\bf \system} & {\bf RCM} & {\bf Trueskill} & {\bf PageRank} & {\bf HITS} & {\bf \system}  \\
\hline
 
\multirow{3}{*}{SO1} & RCM & {\bf {\color{red}55.6}} & 54.0 & 62.8& 52.8&33.3 &{\bf {\color{red}55.7}} & 54.1& 62.8& 52.9&48.1 \\ 
& Trueskill &55.6 & {\bf 54.0}  & 62.8 & 52.8 & 33.3 &55.7 & {\bf 54.1} & 62.8 & 52.9 & 48.1\\ 
& PageRank &42.2 & 56.1 & {\bf 50.3} & 53.5 & 38.1 &42.3 & 56.2 & {\bf 50.5} & 53.5 & 50.5 \\ 
& \system & 46.5& 66.6 & 62.1 & {\bf 49.4}  & {\color{blue}{\bf 71.6}} &46.6 & 66.9 & 62.4 & {\bf 49.4} & {\bf {\color{blue}71.7}} \\\hline

\multirow{3}{*}{SO2} & RCM & {\bf {\color{red}57.6}} &52.8 &51.5 &49.2 &32.4 &{\bf  {\color{red}57.3}}& 52.9&51.6 &49.3 &50.0 \\
& Trueskill &57.6 & {\bf 52.8}  & 51.5 & 49.2 & 32.4 &57.6 & {\bf 52.9} & 51.6 & 49.3 & 50.1\\ 
& PageRank & 46.9& 49.3&{\bf 49.3} & 49.6& 34.2 &47.0 &49.4 &{\bf 49.4} &49.6 & 48.7 \\ 
& \system & 46.9 & 56.1 & 57.2 & {\bf 54.6} & {\color{blue}{\bf 70.5 }} & 46.9& 56.2 & 57.3 &{\bf 54.6} & {\bf {\color{blue}70.5}}\\ \hline

\multirow{3}{*}{SO3} & RCM & {\bf 50.7} &52.2 &54.6 &48.8 &36.9 & {\bf 50.7}&53.2 &55.4 &50.1 &50.0 \\ 
& Trueskill & 50.7& {\bf {\color{red}52.2}} & 54.6 & 48.8 & 36.9 &50.7 & {\bf {\color{red}53.2}} & 55.4 & 50.1 & 50.0\\ 
& PageRank & 51.71& 51.1 & {\bf 50.3} & 44.9 & 36.9 & 52.2& 51.4 & {\bf 51.0} & 46.0 & 50.0\\ 
& \system &54.1 & 60.7 &58.9 &{\bf 49.3} &  {\color{blue}{\bf 76.3}} &54.7 & 62.2 & 60.7& {\bf 50.8} & {\bf {\color{blue}77.1}}\\ \hline

\multirow{3}{*}{MSE} & RCM & {\bf 55.3} &58.6 &55.9 &52.6 &32.1 &{\bf 55.3} &58.6 &55.9 &52.6 &48.7 \\ 
& Trueskill & 55.3& {\bf 58.6} & 55.9 & 52.6 & 32.2 &55.3 & {\bf 58.6} & 55.9 & 52.6 & 48.7\\ 
& PageRank &49.3 & 51.7 & {\bf 53.6} & 56.2 & 34.6 & 49.4& 51.8 & {\bf 53.6} & 56.2 & 49.1 \\ 
& \system &45.6 & 54.8 &61.6 & {\bf {\color{red}58.9}} & {\color{blue}{\bf 71.8 }} & 45.7&54.8&61.6&{\bf{\color{red} 59.0}}& {\bf {\color{blue}72.3}}\\\hline 

\end{tabular}
\newline
\caption{Accuracy (in terms of F\textsubscript{1} and AUC) of the competing methods on four different datasets. \system~is run with its default configuration. Boldface numbers are the accuracy of the baseline method using the configuration reported in the original paper. Blue (red) numbers are the accuracies of the best performing (second-ranked) method. We also measure the accuracy of each competing method using the network suggested by other methods and notice that most of the methods perform better if \system's network is fed into their model (which indeed shows the superiority of our network construction mechanism).}
\label{tab:result}
\end{table*}

\subsection{Comparative Evaluation}
Table \ref{tab:result} shows the accuracy of all the competing methods across different datasets (see the boldface values only). Irrespective of the dataset and the evaluation measures, \system~outperforms all the baselines by a significant margin. \system~acheicves a consistent performance of more than 70\% F\textsubscript{1} score and AUC across different datasets. However, none of the single baseline stands as the best baseline consistently across all the datasets (see the red numbers in Table \ref{tab:result}). \system~beats the best baseline by 28.77\% (28.72\%), 22.79\% (23.03\%), 46.16\% (44.92\%) and 21.90\% (22.54\%) higher F\textsubscript{1} score (AUC) for SO1, SO2, SO3 and MSE datasets respectively.  

Since all the baselines first create a network and run their difficulty estimation modules on the network, one may argue how efficient our network construction mechanism is as compared to other mechanisms. To check this, we further consider each such network (obtained from each competing method) separately, run the difficulty estimation module suggested by each competing method and report the accuracy in Table \ref{tab:result}. For instance, the value reported in second row and first column ($55.6$) indicates the accuracy of the RCM on the network constructed by Trueskill. We once again notice that in most cases, the competing methods perform the best with \system's network. This indeed shows the superiority of our network constriction mechanism.  

\subsection{Feature and Hypothesis Importance}
We also measure the importance of each feature  for \system~with default configuration. For this, we drop each feature in isolation and measure the accuracy of \system. Figure \ref{fig:factor_imp}(a) shows that the maximum decrease in accuracy (27.63\% decrease in F\textsubscript{1}) is observed when we drop leader follower ranking (F1 and F2), followed by PageRank (F3 and F4) and degree (F5 and F6). However, there is no increase in accuracy if we drop any feature, indicating that all features should be considered for this task.

One would further argue on the importance of keeping all hypotheses (all edge type). For this, we conduce similar type of experiment -- we drop each edge type (by ignoring its corresponding hypothesis), reconstruct the network and measure the accuracy of \system. Figure \ref{fig:factor_imp}(b) shows  similar trend -- none of the dropping of edges increases the accuracy, which indicates that all types of edges are important. However, type 2 edges sees to be more important (dropping which reduces 26.31\% F\textsubscript{1} score), followed by type 1 and type 3 edges.

\begin{figure}[t!]
    \centering
           \includegraphics[width=\columnwidth]{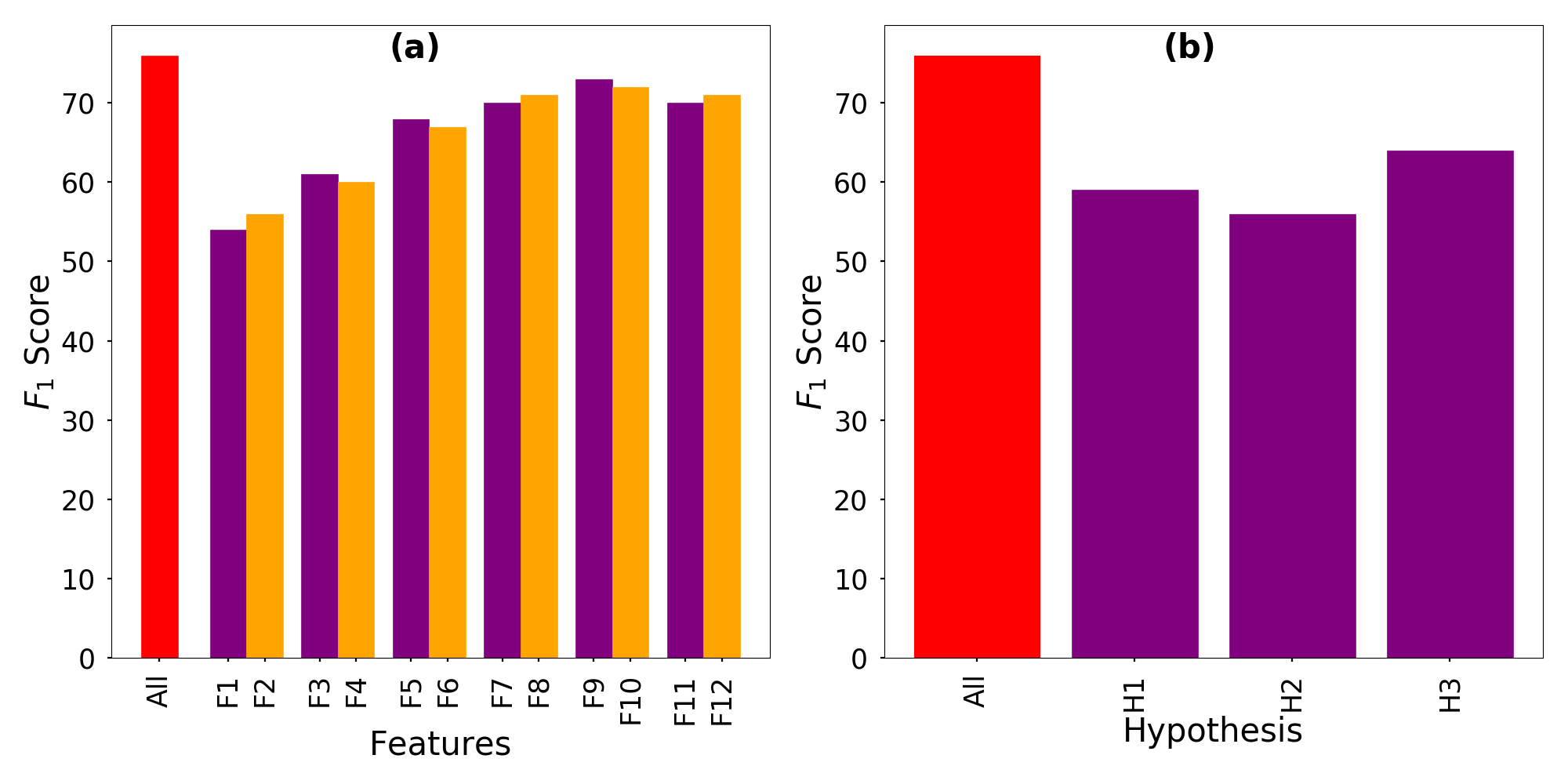}     
           \caption{Importance of (a) features and (b) hypotheses based on F\textsubscript{1} score for SO3 dataset. For better comparison, we all show the results with all features and all hypotheses (All). {\color{red} add x axis}
              \label{fig:factor_imp}}
\end{figure}

\subsection{Robustness under Noise}
We further test the robustness of \system~under network noise. We employ two types of noise into the network: (i) {\bf Noise 1}, where we keep inserting $x\%$ of the existing edges randomly into the network, thus increasing the size of the training set, and (ii) {\bf Noise 2}, where we first remove $x\%$ of the existing edges and randomly insert $x\%$ of the edges into the network, thus the size of the training set remaining same after injecting the noise. We vary $x$ from $5$ to $20$ (with the increase of $5$). We hypothesis that a robust method should be able to tolerate the effect of noise up to a certain level, after which its performance should start deteriorating. 
Figure \ref{fig:noise} shows the effect of both types of noises on the performance of \system. \system~seems to tolerate 5\% of noise 1 (??\% of performance degradation) and 10\% of noise 2 (??\% of performance degradation), after which the performance decreases significantly. With 15\% of noise 1\todo{need Trueskill result} and 10\% of noise 2, the accuracy of \system~deteriorates up to ??\% and ??\%. Therefore, we can conclude that \system is appropriately sensitive to network noise. Note that as expected, the effect of noise 1 seems to be higher than the effect of noise 2.

\begin{figure}[!h]

    \centering
           \includegraphics[width=\columnwidth]{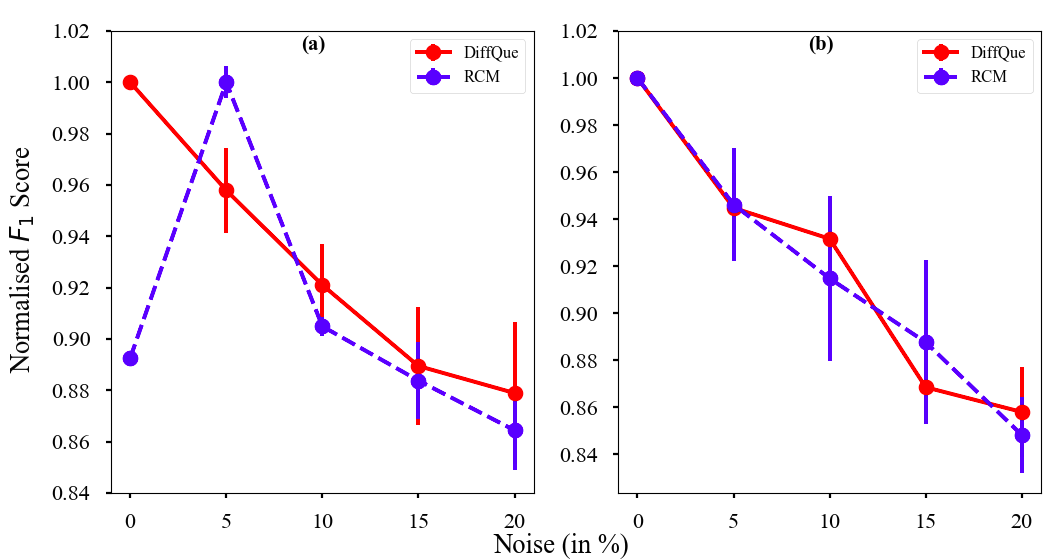}     
           \vspace{-5mm}
           \caption{Change in the accuracy (F\textsubscript{1} score) of \system~with the increase of network noise on SO3 dataset (similar pattern is observed for other datasets){\color{red}add RCM }.}
               \label{fig:noise}
\end{figure}

\subsection{Capability of Domain Adaptation}
Another signature of a robust system is that it should be capable of adopting different domains, i.e., it should be able to get trained on one dataset and perform on another dataset. To verify this signature, we train \system~ and Trueskill on one dataset and test them on another dataset.  Table \ref{tab:domain} shows that \todo{need to write} 

\begin{table}
\centering
\scalebox{0.85}{
\begin{tabular}{|c|c|c|c|c|c|}
\hline
\multicolumn{6}{|c|}{{\bf Testing}}\\\hline
& & {\bf SO1} & {\bf SO2} & {\bf SO3} & {\bf MSE} \\\hline
\parbox[t]{2mm}{\multirow{4}{*}{\rotatebox[origin=c]{90}{{\bf Training}}}}& {\bf SO1}  & 71.65 (xx.xx) & 66.99 (xx.xx) & 76.68 (xx.xx) & 71.49 (xx.xx)\\\cline{2-6}
& {\bf SO2}  & 70.21 (xx.xx) & 70.53 (xx.xx) & 77.24 (xx.xx) & 67.60 (xx.xx) \\\cline{2-6}
&{\bf SO3}  & 70.27 (xx.xx) & 66.65 (xx.xx) & 76.39 (xx.xx) & 71.22 (xx.xx) \\\cline{2-6}
&{\bf MSE}  & 70.98 (xx.xx) & 66.99 (xx.xx) & 76.36 (xx.xx) & 71.84 (xx.xx) \\\hline
\end{tabular}}
\caption{F\textsubscript{1} score of \system~ and Trueskill (within parenthesis) for different combination of training and testing sets.{\color{red} fill up the blanks} }\label{tab:domain}
\end{table}

\begin{table}
\centering
\scalebox{0.85}{
\begin{tabular}{|c|c|c|c|c|c|}
\hline
\multicolumn{6}{|c|}{{\bf Testing}}\\\hline
& & {\bf SO1} & {\bf SO2} & {\bf SO3} & {\bf MSE} \\\hline
\parbox[t]{2mm}{\multirow{4}{*}{\rotatebox[origin=c]{90}{{\bf Training}}}}&
{\bf SO1}              & 54    & 52.3  & 52.35 & 58.97 \\\cline{2-6}
&{\bf SO2}              & 52.48 & 52.8  & 52.27 & 59.32 \\\cline{2-6}
&{\bf SO3}              & 52.3  & 52.49 & 52.2  & 57.65 \\\cline{2-6}
&{\bf MSE}              & 51.64 & 52.49 & 53.02& 58.6\\\hline
\end{tabular}}
\caption{F\textsubscript{1} score of TrueSkill  for different combination of training and testing sets.}\label{tab:domain}
\vspace{-5mm}
\end{table}

\begin{table}
\centering
\scalebox{0.85}{
\begin{tabular}{|c|c|c|c|c|c|}
\hline
\multicolumn{6}{|c|}{{\bf Testing}}\\\hline
& & {\bf SO1} & {\bf SO2} & {\bf SO3} & {\bf MSE} \\\hline
\parbox[t]{2mm}{\multirow{4}{*}{\rotatebox[origin=c]{90}{{\bf Training}}}}&
{\bf SO1}              & 50.3  & 49.01 & 43.98 & 52.44 \\\cline{2-6}
&{\bf SO2}              & 49.82 & 49.3  & 46.75 & 53.05 \\\cline{2-6}
&{\bf SO3}              & 49.51 & 49.41 & 50.3  & 53.77 \\\cline{2-6}
&{\bf MSE}              & 52.18 & 48.22 & 48.04 &53.6\\\hline
\end{tabular}} 
\caption{F\textsubscript{1} score of PageRank  for different combination of training and testing sets.}\label{tab:domain}
\vspace{-5mm}
\end{table}

\begin{table}
\centering
\scalebox{0.85}{
\begin{tabular}{|c|c|c|c|c|c|}
\hline
\multicolumn{6}{|c|}{{\bf Testing}}\\\hline
& & {\bf SO1} & {\bf SO2} & {\bf SO3} & {\bf MSE} \\\hline
\parbox[t]{2mm}{\multirow{4}{*}{\rotatebox[origin=c]{90}{{\bf Training}}}}&
{\bf SO1}              & 49.4  & 54.63 & 49.39 & 57.84 \\\cline{2-6}
&{\bf SO2}              & 48.1  & 54.6  & 48.97 & 61.47 \\\cline{2-6}
&{\bf SO3}              & 52.54 & 51.34 & 49.3  & 58.42 \\\cline{2-6}
&{\bf MSE}              & 48.92 & 53.58 & 45.53 & 58.9\\\hline
\end{tabular}}
\caption{F\textsubscript{1} score of HITS  for different combination of training and testing sets.}\label{tab:domain}
\vspace{-5mm}
\end{table}

\begin{figure}[!h]
\vspace{-1mm}
    \centering
           \includegraphics[width=\columnwidth]{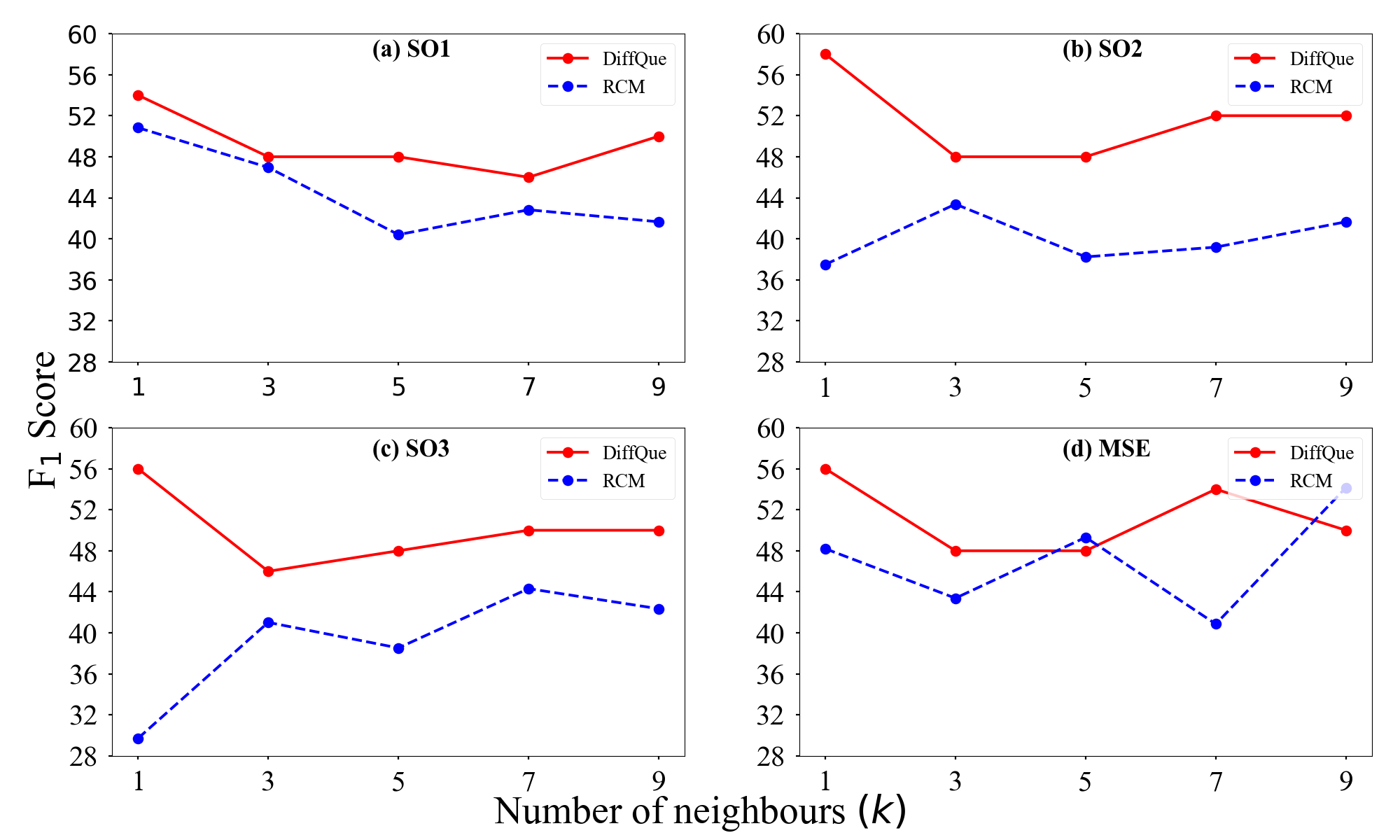}     
           \caption{Different value of neighbours and the normalised F\textsubscript{1} score obtained{\color{red} why normalized F1}}
               \label{fig:coldstart}
               \vspace{-1mm}
\end{figure}

\subsection{Handling Cold Start Problem}
Most of the previous research \cite{chinese-paper,msr-paper} deal with well-resolved questions which have received enough attention in terms of the number of answers, upvotes etc. However, they suffer from the `cold start' problem -- they are inefficient to handle newly posted questions with no answer and/or the users who posted the question are new to the community.  In many real-world applications such as question routing and incentive mechanism design, however, it is usually required that the difficulty level of a question is known instantly after it is posted \cite{regularised}. 

\system~overcomes the cold start problem by exploiting the textual description of questions. In our network construction, a question which has been asked by a completely new user forms an isolated node; therefore none of the network related features (such as F1-F6) cannot be computed for that question. Moreover, if it does not receive any accepted answer, other features (F7-F12) cannot be measured. We call those questions ``brand-new'' for which none of the features can be computed, leading to extreme level of cold start. Given a question pair  ($Q_1, Q_2$) under inspection,  there can be two types of cold start scenarios: (i) both $Q_1$ and $Q_2$ are brand new, (ii) either $Q_1$ or $Q_2$ is brand-new. To handle the former, we run  Doc2Vec \cite{Le:2014}, a standard embedding technique on textual description of all the questions and return $k$ most similar questions (based on cosine similarity) for each of $Q_1$ and $Q_2$, which are not brand-new. For $Q_1$ ({\em resp.} $Q_2$), the $k$ most similar questions are denoted as $\{Q^i_1\}_{i=1}^k$ ({\em resp.} $\{Q^i_2\}_{j=1}^k$). We assume each such set as the representative of the corresponding brand-new question. We then run \system~to measure the relative difficulty of each pair (one taken from $\{Q^i_1\}$ and other take from $\{Q^i_2\}$). Finally, we consider that brand-new question to be difficult whose corresponding neighboring set is marked as difficult maximum times by \system. The latter cold start scenario is handle in a similar way. Let us assume that $Q_1$ is old and $Q_2$ is brand-new. For $Q_2$, we first extract $k$ most similar questions $\{Q^i_2\}$ using the method discussed above. We then run \system~ to measure the relative difficulty between $Q_1$ and each of $\{Q^i_2\}$, and make the one as difficult which wins the most.          
 
To test the efficiency of our cold start module, we take $50$ annotated pairs (edges) randomly for each dataset and remove those edges (and their associated nodes) from the network. These pairs form the test samples for the cold start problem.  Figure \ref{fig:coldstart} shows the accuracy of \system~and the best baseline (Trueskill) with the increase of $k$ to handle cold start problem.


\bibliographystyle{ACM-Reference-Format}
\bibliography{sample-bibliography} 
\fi

\end{document}